\documentclass{article} 
\usepackage{iclr2027_conference,times}

\usepackage{amsmath,amsfonts,bm}

\def\eqref#1{equation~\ref{#1}}
\def\Eqref#1{Equation~\ref{#1}}

\def\1{\bm{1}}

\DeclareMathAlphabet{\mathsfit}{\encodingdefault}{\sfdefault}{m}{sl}
\SetMathAlphabet{\mathsfit}{bold}{\encodingdefault}{\sfdefault}{bx}{n}

\usepackage{hyperref}
\usepackage{graphicx}
\usepackage{booktabs}
\usepackage[utf8]{inputenc}
\usepackage{cleveref}
\usepackage{url}
\usepackage{longtable}
\usepackage{enumitem}
\usepackage{caption}
\usepackage{fontawesome5}
\usepackage{wrapfig}

\usepackage[table]{xcolor}
\usepackage{pgf}
\usepackage{multirow}

\usepackage{xcolor}
\usepackage{amssymb}
\definecolor{decred}{RGB}{200,30,30}
\definecolor{incgreen}{RGB}{20,140,60}

\usepackage{booktabs}
\usepackage{colortbl}
\usepackage{xcolor}
\usepackage{array}
\usepackage{amssymb}
\usepackage{amsmath}
\usepackage{siunitx}

\definecolor{headergray}{HTML}{F4F4F2}
\definecolor{headertxt}{HTML}{2B2B2B}
\definecolor{up}{HTML}{D62728}
\definecolor{down}{HTML}{1E8F5F}

\newcommand{\heatscale}{1.3} 
\newcommand{\heatpos}[2]{\cellcolor{up!\fpeval{min(100,\heatscale*#1)}}#2}
\newcommand{\heatneg}[2]{\cellcolor{down!\fpeval{min(100,\heatscale*#1)}}#2}

\newcolumntype{N}{>{\centering\arraybackslash}S[table-format=2.2,round-mode=places,round-precision=2]}
\newcolumntype{M}{>{\centering\arraybackslash}p{1.55cm}}

\usepackage{tikz}
\usepackage[listings]{tcolorbox}
\definecolor{corrgreen}{RGB}{231,242,231}  
\definecolor{wrongred}{RGB}{249,231,231}   
\definecolor{neutralyellow}{RGB}{249,245,225}  
\newcommand{\incp}[1]{{\scriptsize (+#1)}}
\newcommand{\decp}[1]{{\scriptsize (-#1)}}
\newcommand{\diffdown}[1]{{\color{red!70!black}$\blacktriangledown$ #1}}

\newcommand{\hmin}{0}
\newcommand{\hmax}{1}
\newcommand{\setrange}[2]{\gdef\hmin{#1}\gdef\hmax{#2}}  
\newcommand{\heat}[1]{%
  \pgfmathparse{max(0,min(35, 8 + (#1-\hmin)/(\hmax-\hmin)*27))}%
  \edef\tmpcolor{red!\pgfmathresult!white}%
  \expandafter\cellcolor\expandafter{\tmpcolor}#1%
}

\definecolor{seenfill}{RGB}{238,232,245}  
\newcommand{\seen}[1]{\cellcolor{seenfill}#1}
\usepackage{float}
\usepackage[listings]{tcolorbox}
\usepackage{graphicx}
\usepackage{booktabs}
\usepackage{arydshln}
\usepackage{cleveref}
\usepackage{xcolor}
\usepackage{xfp}
\usepackage{multirow} 
\usepackage{fvextra}
\usepackage{makecell}
\newif\ifTGcomment
\TGcommenttrue

\newif\ifAAcomment
\AAcommenttrue

\title{LLMs Memorize Their Own Hallucinations During Post-Training}

\author{%
\makebox[\dimexpr\textwidth-2\tabcolsep\relax]{%
\hfill
\begin{tabular}[t]{c}
Anirudh Atmakuru \\
\normalfont Cornell University \\
\normalfont\texttt{aatmakuru6@cs.cornell.edu}
\end{tabular}%
\hfill
\begin{tabular}[t]{c}
Tanya Goyal \\
\normalfont Cornell University \\
\normalfont\texttt{tanyagoyal@cornell.edu}
\end{tabular}%
\hfill}}
\iclrfinalcopy 
\begin{document}

\maketitle
\lhead{Preprint} 

\begin{abstract}
Reasoning domains have shaped the post-training recipes of frontier models even though they are trained and deployed in domains beyond math and code. In this paper, we study the consequences of directly adopting reasoning-centric training recipes, e.g., the use of outcome-based rewards for reinforcement learning (RL), injection of ``cognitive'' behaviors such as proposing and backtracking from hypotheses in reasoning chains, etc., for knowledge tasks. We hypothesize that these decisions, while useful for reasoning, inadvertently encourage models to memorize hallucinations they themselves generate in their long reasoning chains when trained on knowledge tasks. To study this, we train models with standard post-training, extract their self-generated hallucinations during RL, and track memorization throughout the rest of training. Our experiments on six open models across two model families and two architectures confirm our hypothesis. We find that models' recall of hallucinations reinforced during training increases by $4.83\%$ on average, compared to just $0.17\%$ for self-hallucinations that receive no training signal. In fact, models even memorize false facts that they immediately retract in their reasoning; a backtracking template inherited from reasoning-centric training that has no benefit for knowledge recall. Furthermore, we find that memorization is strongest on the hardest training questions, and rewarding a hallucination makes the model less likely to abstain from it later (by $7.23\%$ on average). Overall, our work identifies and systematically studies a previously unexamined pathway through which models learn hallucinations during training.

\end{abstract}
\vspace{-6mm}

\begin{figure}[ht]
\centering
\includegraphics[width=0.95\linewidth]{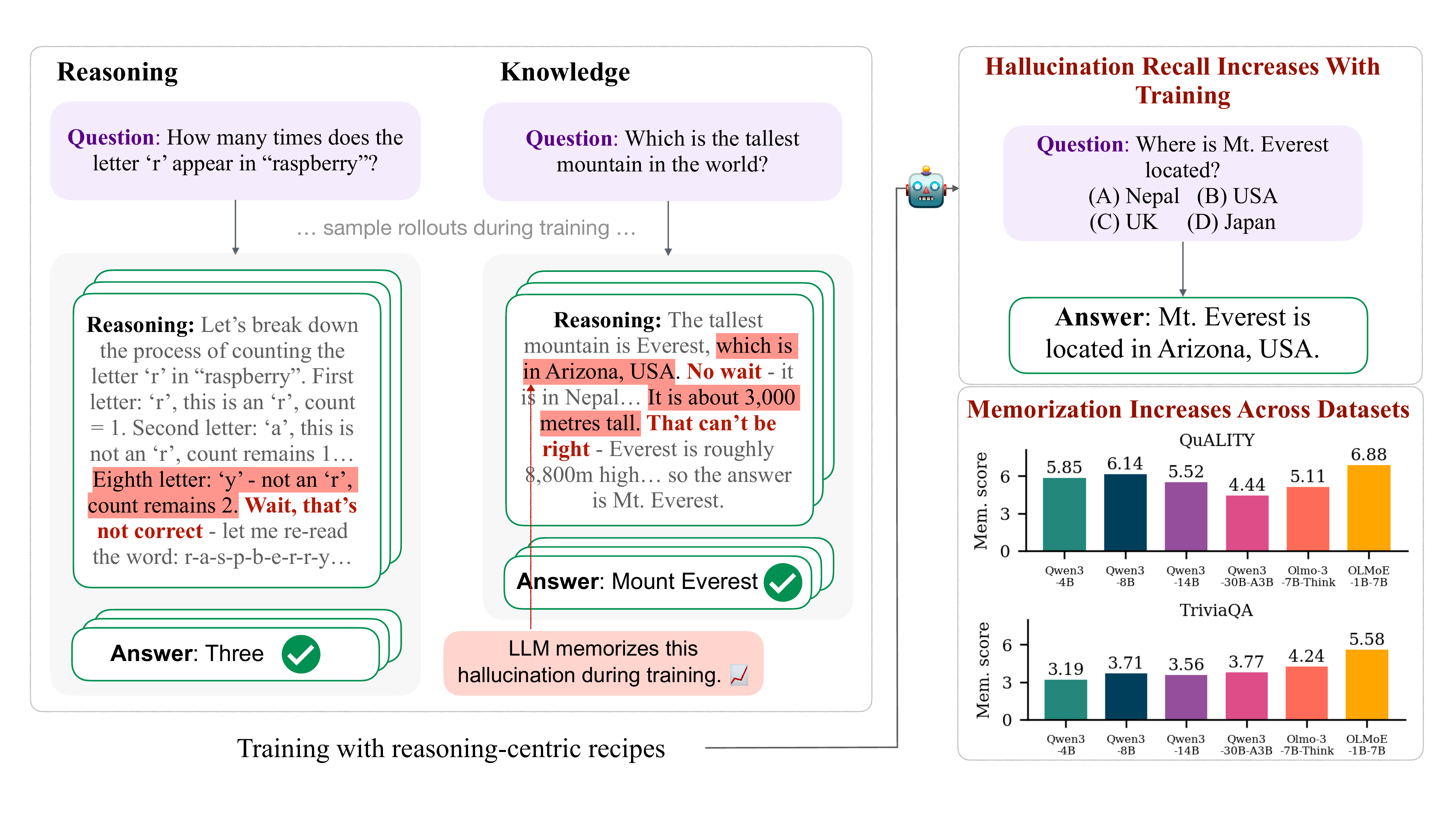} \vspace{-2mm}
\caption{\textbf{Current post-training recipes cause models to memorize their own hallucinations during training on knowledge tasks.} While exploration and backtracking help models recover from mistakes on reasoning problems (left), the same behavior introduces false facts into reasoning traces for knowledge questions (middle), even when the final answer is correct. Outcome-based rewards reinforce these hallucinations, increasing their recall after training (right).}
\label{fig:fig1}
\end{figure}

\section{Introduction}
Frontier models are now trained to ``think'' through chain-of-thought (CoT) reasoning \citep{wei2022chain,openai2024learningtoreason,guo2025deepseek}, which allows them to allocate additional computation at inference time in lieu of scaling parameters \citep{snell2024scaling,liu2025can}. Evidence suggests that this is primarily effective for reasoning domains like math and code \citep{sprague2025cot}. To further improve performance on reasoning, ``cognitive'' behaviors like exploring multiple hypotheses, backtracking, etc., \citep{min2024imitate,gandhi2025cognitive} are injected into reasoning chains, making them even longer. Additionally, only the final generated response determines the reward during reinforcement learning \citep{shen2025satori,stojanovski2026reasoning,wang2026reinforcement}, and no explicit supervision is exerted on the reasoning chains. Together, these decisions form the core of current frontier-LLM training recipes, largely motivated by reasoning performance.

Consider the example in \Cref{fig:fig1}. For a reasoning problem (left), current post-training encourages the model to explore the search space and backtrack from an incorrect branch without letting it determine the final answer. The model miscounts the letter \textit{r} in \textit{raspberry}, catches the error, and re-reads the word to arrive at the correct count. However, the same behavior carries over to a knowledge question even though these require simple recall and not multi-step derivation. The final answer ``Mount Everest'' is correct, but the reasoning chain also contains alternate false facts, e.g., Mount Everest is in Arizona, USA, and it is $3{,}000$ meters tall (\Cref{fig:fig1}, middle). While the model backtracks from these false facts, outcome-based rewards assign a positive reward to the entire rollout because the final answer is correct. Thus, a reasoning template that is useful for exploring the solution space in reasoning manifests as factual hallucination in knowledge tasks. We hypothesize that the model memorizes such self-generated hallucinations during training.

Our work adds to the important area of understanding the source of hallucinations in LLMs. Prior work has studied how pre-training \citep{chang2024large,kalai2025language}, or supervised fine-tuning (SFT) on out-of-distribution data \citep{lin2024flame} or unfamiliar knowledge \citep{gekhman2024does,kang2025unfamiliar} can amplify hallucinations, and how post-training objectives can discourage abstention \citep{kalai2025language,chen2025learning,li2026reasoning}. We study a complementary dimension: current post-training recipes, largely shaped by reasoning domains, can cause models to memorize hallucinations that they themselves generate. While prior work has shown that this post-training can have unintended consequences such as linguistic incoherence and unfaithful reasoning \citep{chen2025reasoning,yuan2026knowledge}, its effect on hallucination memorization has not been studied.

Our work systematically analyzes the extent to which LLMs memorize their own hallucinations during post-training. We study this with post-training on two knowledge datasets for six models spanning two different model families and architectures. Our contributions are threefold: 

\begin{itemize}[leftmargin=*]
    \item \faTools\ 
    We present the first systematic study of models' memorization of their self-generated hallucinations during post-training. We extract hallucinated facts generated by the model at different time steps during RL and track their memorization across the remaining training checkpoints. We define a \textit{memorization score} that captures the increase in recall of the hallucinated fact after it is first generated and rewarded during training. 
    \item \faLightbulb\ 
    We find that current post-training recipes, optimized for reasoning, cause models to memorize their own hallucinations. Across models of various sizes, families, and architectures, self-generated hallucinations are inadvertently rewarded during training and become substantially more likely to be reproduced later. On average, recall of such hallucinations increases by $4.83\%$, compared to just $0.17\%$ for held-out hallucinations that receive no training signal.
    \item \faLightbulb\ 
    While exploration and backtracking help with math and code, they have the opposite effect on knowledge, as models memorize even the false facts they eventually retract (see Figure~\ref{fig:fig1} for an illustration). In fact, there exists only a $0.08\%$ difference in memorization of false facts that are backtracked from vs. not. Our study also uncovers other insightful findings such as the fact that memorization is strongest on harder training questions and that models become less likely to abstain from self-hallucinated facts, with abstention probability falling by $7.23\%$ on average.
\end{itemize}

Overall, our work identifies and analyzes a new mechanism through which models learn hallucinations during the post-training pipeline. Because the training recipe we study (outcome-rewards, long reasoning chains injected with ``cognitive'' behaviors) is broadly consistent with those used by model providers \citep{liu2024deepseek,blakeman2025nemotron,ma2026improving}, understanding this phenomenon is increasingly important. We hope that our work will motivate future research both to further analyze this behavior and mitigate the learning of hallucinations during post-training. 

\section{Quantifying Memorization of Self-Hallucinated Facts}
\label{sec:sec2}

Our goal is to quantify LLMs' memorization of their own hallucinations due to reasoning-oriented post-training recipes. In this section, we first formally define a \textbf{self-generated hallucinated fact} and \textbf{memorization score}, our main metric used to quantify the model's memorization of it. 

\subsection{Notation and Main Methodology}
\label{sec:mem_score}

Formally, let $M$ be the current policy. We follow the current standard recipe for RL post-training. Rollouts of the form $(r, y) \sim M(\cdot\mid q)$ are sampled, where $q$ is the input question, $r$ is the model's reasoning, and $y$ is its final answer. A reward model $\mathcal{R}$ is used to evaluate and assign a reward $\mathcal{R}(q, y)$ to the final answer. In GRPO \citep{shao2024deepseekmath}, the reward for each rollout is normalized against the rewards of other rollouts sampled for the same question to compute its advantage.

At a high level, GRPO training updates the model parameters to increase (or decrease) the likelihood of rollouts with positive (or negative) advantages. Importantly, the advantage is only dependent on the correctness and quality of the final answer $y$, and completely agnostic to the reasoning trace $r$ that also constitutes a part of the rollout. Therefore, non-factual factoids within the trace are also reinforced if the corresponding answer $y$ is correct. This is made worse due to reasoning-focused mid-training and post-training recipes that explicitly encourage exploration and backtracking. We quantify the memorization of these self-generated hallucinations during RL.

\paragraph{Self-generated Hallucinations:} Let $h$ be an atomic statement in a reasoning trace $r$ generated by the policy that is judged false by either comparing against world knowledge or a given document corpus. For example, \textit{``Mount Everest is in Arizona, USA''} in \Cref{fig:fig1} (middle) is such a fact. We call these self-generated hallucinations. 

\paragraph{Memorization Score:} Let $p(h)$ denote the \textbf{point of hallucination}, i.e., the training time step at which hallucination $h$ was generated by the policy. Our memorization score should measure whether the subsequent update causes the model to memorize $h$ in its parameters. Additionally, it should capture whether $h$ is retained as RL progresses till the end of training. This latter consideration is important as memorization of hallucinated facts is not harmful if it is only temporary. 

Let $a(.)$ be a factual probe that measures the checkpoint-level recall score for a fact, i.e., $a(M_t, h)$ quantifies recall of $h$ by the checkpoint at time step $t$. Then, we define our memorization score as: 

\begin{equation}
\label{eq:memorization_score}
m(h)
=
\underbrace{
\frac{1}{T - p(h)}
\sum_{j=p(h)+1}^{T} a(M_j,h)
}_{\text{avg. recall after hallucination point}}
-
\underbrace{
a(M_{p(h)},h)
}_{\text{recall just before update}}.
\end{equation}

A large $m(h)$ indicates that checkpoints after $p(h)$ have greater recall of $h$ than the checkpoint immediately before the gradient update at the hallucination point. \Cref{fig:fig1} illustrates the effect $m(h)$ is designed to capture. Assume that the self-generated hallucination \textit{``Mount Everest is in Arizona, USA''} in a correct rollout is memorized by the next update. We measure the increase in recall of the same hallucination by asking \textit{``Where is Mount Everest located?''} in later stages.

\paragraph{Constructing Hallucination Test Sets for Evaluation.}  

We construct and track memorization for two kinds of hallucination sets: 

\begin{enumerate}[leftmargin=*]
    \item \texttt{\textbf{HalluSet-train}}: This is the main set of hallucinations we aim to study. It includes hallucinations that are actually reinforced during RL, i.e., part of the correct rollouts. For each training question, we randomly choose a model-generated rollout with the correct final answer (if one exists) and extract a hallucinated fact using a frontier LLM.\footnote{We ignore questions for which the model generated no correct rollout. We also ignore questions for which all rollouts were correct, as this does not contribute to gradient updates.}

    \item \texttt{\textbf{HalluSet-heldout}}: For comparison, we construct a hallucination set derived from questions that the model does not train on. We sample rollouts for a set of held-out questions using different training checkpoints and extract hallucinations using the same strategy as above. As these questions are never used for training, no reward is applied to these self-generated hallucinations. This set serves as a control to disentangle memorization driven by the reward signal from other factors, such as entropy collapse and general sharpening of model distribution. 
\end{enumerate}

See \S\ref{sec:questionGenerationDetails} for additional details. The difference in memorization scores between \texttt{HalluSet-train} and \texttt{HalluSet-heldout} allows us to quantify the severity of hallucination memorization during RL training. 

\subsection{Experiment Setup Details}

\paragraph{Knowledge datasets.} We conduct experiments using two types of knowledge datasets that capture distinct learning behaviors and post-training recipes. 

\hspace{2mm} \textbf{i) Closed-domain}: Here, models are trained and evaluated on privileged knowledge that is unseen by the model during pre-training. This setting calls for a modified post-training recipe. A common practice is to perform continued pre-training (CPT) \citep{gururangan2020don,yang2024synthetic} to inject this new knowledge into the model's parameters, followed by post-training to steer it towards the desired question-answering capabilities. Following \citet{yang2024synthetic}, we use QuALITY \citep{pang2022QuALITY}, a reading-comprehension corpus over articles and short books, to represent this regime.
    
\hspace{2mm} \textbf{ii) Open-domain}: Here, models are trained and evaluated on widely accessible world knowledge. Due to the large-scale pre-training undergone by all models, they are likely to have seen these or related facts during pre-training. Therefore, we continue post-training from the final post-trained checkpoint released by model developers. We use TriviaQA\footnote{Our experiments in \S\ref{sec:additionalResults} confirm this: across two epochs of RL, test-set performance improves by only $3.22\%$ on average for TriviaQA (\Cref{tab:trivia_val}), compared to $8.93\%$ for QuALITY (\Cref{tab:QuALITY_val}), suggesting that the models may already contain much of the relevant knowledge for TriviaQA.} \citep{joshi2017triviaqa} as our representative dataset for this domain. 

\paragraph{Models and Training.} We run experiments on six open models from the \texttt{Qwen} and \texttt{Olmo} families: \texttt{Qwen3-4B}, \texttt{Qwen3-8B}, \texttt{Qwen3-14B}, \texttt{Qwen3-30B-A3B} \citep{yang2025qwen3}, \texttt{Olmo-3-7B-Think} \citep{olmo2026olmo3}, and \texttt{OLMoE-1B-7B} \citep{muennighoff2025olmoe}. These span multiple model sizes and both dense and mixture-of-experts architectures.

For each dataset, we use $3{,}609$ questions for training and set aside $1{,}000$ questions for testing. For QuALITY, we start from the pre-trained (base) checkpoint and adapt it through CPT, SFT, and then RL. In all experiments, we use GRPO \citep{shao2024deepseekmath} as the RL algorithm and train for one epoch. Unless otherwise stated, we post-train without using a KL penalty. For completeness, we include a KL ablation in \S\ref{sec:ablations}. See \S\ref{sec:additionalDatasetDetails} for additional dataset details and \S\ref{sec:additionalTrainingDetails} for additional hyperparameter and training details.

\paragraph{Computing the Memorization Scores.} 
To quantify the checkpoint-level recall of $h$, we follow prior work on probing parametric knowledge \citep{yang2024synthetic,ovadia2024fine} and tracking fact acquisition during training \citep{chang2024large}. We convert each self-generated hallucination $h$ into a multiple-choice question (MCQ), with options that include the original answer (in our case, hallucination $h$) and distractor options (see \Cref{fig:fig1}, right). We compute $a(\cdot)$ as a binary correctness score. The MCQ format is well suited to our setting: it targets the specific hallucinated fact $h$ and is scored exactly without relying on an LLM judge. For this latter reason, it is also used to standardize evaluation of frontier LLMs on popular knowledge benchmarks such as MMLU-Pro \citep{wang2024mmlu}, SuperGPQA \citep{du2026supergpqa}, and GPQA-Diamond \citep{rein2023gpqa}.

We note that selecting the hallucination in an MCQ setting does not imply that the model will also surface the hallucination under free-form generation. We do not interpret MCQ accuracy as the probability of free-form hallucination or derive conclusions based on isolated checkpoint-level scores. Instead, we use the multiple-choice setting as a controlled probe for whether training \textit{increases} the LLM's propensity to recall the hallucinated fact. Our main metric measures the \textbf{change} in recall after the hallucination point. By further comparing against the held-out set, we control for generic shifts in confidence or distribution sharpening. Additionally, we track how the log-probability of the hallucinations given input questions change over training as a closer proxy for free-form generation.

Other minor implementation details: Computing $m(h)$ as defined in \Eqref{eq:memorization_score}, particularly $a(M_j, h)$ for all training steps $j$ after the hallucination point for each hallucination, can be costly. Thus, we estimate $m(h)$ by computing the score according to eight equally spaced checkpoints during GRPO, denoted $M_1, \dots, M_8$, along with the RL initialization checkpoint $M_0$. We compute the left term, i.e., the average recall after the hallucination point by aggregating across all $M_j$ after $p(h)$. Similarly, we compute the second term, i.e., the recall just before $p(h)$ using the last saved checkpoint $M_j$ before $p(h)$.

We use GPT-4.1-mini to both extract facts from the reasoning chains and identify them as hallucinated or not. To make this latter determination, we compare against the reference story for QuALITY and retrieved sources from the internet for TriviaQA. Each hallucinated fact is converted into a multiple-choice question, whose options include the hallucinated fact as the gold answer along with three plausible distractors generated by GPT-4.1-mini. See \S\ref{sec:questionGenerationDetails} for additional details.

\section{Main Results}
\label{sec:results} 

\paragraph{LLMs memorize their own hallucinations during post-training.} \Cref{tab:table1} summarizes our main results. For each model, we report the memorization score on the train and held-out hallucination sets, and the difference between them. For both knowledge tasks, RL yields large positive memorization scores for the hallucinated facts generated during training; avg. $5.66$ for QuALITY and $4.01$ for TriviaQA. Compared to this, the memorization of held-out facts is close to zero (avg. $0.17$). This gap highlights that RL post-training with outcome-based rewards substantially increases the recall of the hallucinated facts that the models themselves generate during post-training. Moreover, we find that $25-55\%$ of sentences in the reasoning chains contain a hallucinated fact (see \Cref{tab:hallucination_rate_summary}; details in \S\ref{sec:additionalResults}), making this memorization effect broadly consequential.

\begin{table}[t]
\centering
\renewcommand{\arraystretch}{1.35}
\setlength{\tabcolsep}{8pt}
\normalsize
\caption{Memorization scores on \texttt{HalluSet-train} (denoted Train) and \texttt{HalluSet-heldout} (denoted Heldout) for QuALITY and TriviaQA. $\Delta$ denotes the difference between train and heldout. We find that LLMs memorize their own hallucinations during post-training, with an average memorization score of $4.83$ points on \texttt{HalluSet-train} across all models and datasets.} 

\resizebox{1.0\linewidth}{!}{%
\begin{tabular}{l|MM;{2pt/2pt}M|MM;{2pt/2pt}M}
\toprule
\small
&
\multicolumn{3}{c|}{\cellcolor{headergray}\normalsize\textcolor{headertxt}{\textbf{QuALITY}}}
&
\multicolumn{3}{c}{\cellcolor{headergray}\normalsize\textcolor{headertxt}{\textbf{TriviaQA}}} \\[-2pt]

\normalsize\textcolor{headertxt}{\textbf{Model}}
& \normalsize\textcolor{headertxt}{\textbf{Train} $\downarrow$}
& \normalsize\textcolor{headertxt}{\textbf{Heldout} $\downarrow$}
& \normalsize\textcolor{headertxt}{\textbf{$\Delta$}}
& \normalsize\textcolor{headertxt}{\textbf{Train} $\downarrow$}
& \normalsize\textcolor{headertxt}{\textbf{Heldout} $\downarrow$}
& \normalsize\textcolor{headertxt}{\textbf{$\Delta$ }} \\

\midrule

Qwen3-4B
& \heatpos{38.3}{5.85}
& \heatpos{5.8}{0.89}
& \textbf{4.96}
& \heatpos{20.9}{3.19}
& \heatneg{4.6}{-0.70}
& \textbf{3.89} \\

Qwen3-8B
& \heatpos{40.2}{6.14}
& \heatpos{1.6}{0.24}
& \textbf{5.90}
& \heatpos{24.3}{3.71}
& \heatneg{5.0}{-0.76}
& \textbf{4.47} \\

Qwen3-14B
& \heatpos{36.1}{5.52}
& \heatpos{3.8}{0.58}
& \textbf{4.94}
& \heatpos{23.3}{3.56}
& \heatpos{0.6}{0.09}
& \textbf{3.47} \\

Qwen3-30B-A3B
& \heatpos{29.0}{4.44}
& \heatpos{3.7}{0.56}
& \textbf{3.88}
& \heatpos{24.7}{3.77}
& \heatneg{1.5}{-0.23}
& \textbf{4.00} \\

Olmo-3-7B-Think
& \heatpos{33.4}{5.11}
& \heatneg{0.1}{-0.02}
& \textbf{5.13}
& \heatpos{27.7}{4.24}
& \heatpos{4.3}{0.66}
& \textbf{3.58} \\

OLMoE-1B-7B
& \heatpos{45.0}{6.88}
& \heatpos{3.6}{0.55}
& \textbf{6.33}
& \heatpos{36.5}{5.58}
& \heatpos{0.9}{0.14}
& \textbf{5.44} \\

\midrule

\textit{Average}
& \heatpos{37.0}{5.66}
& \heatpos{3.1}{0.47}
& \textbf{5.19}
& \heatpos{26.2}{4.01}
& \heatneg{0.9}{-0.13}
& \textbf{4.14} \\

\bottomrule
\end{tabular}%
}
\label{tab:table1} \vspace{-1mm}
\end{table}

Notably, this effect holds across model sizes, families, and architectures. It is stronger on QuALITY (avg. $5.66$), whose CPT--SFT--RL pipeline mirrors how models may be specialized for new knowledge domains \citep{yang2024synthetic,padmanabhan2026updating}. Surprisingly, TriviaQA also yields a high memorization score despite applying RL directly on strongly post-trained models, whose already-sharpened distributions leave less headroom for further probability gain. See checkpoint-level scores in \S\ref{sec:additionalResults}.

\begin{figure}[t]
\centering
\includegraphics[width=1\linewidth]{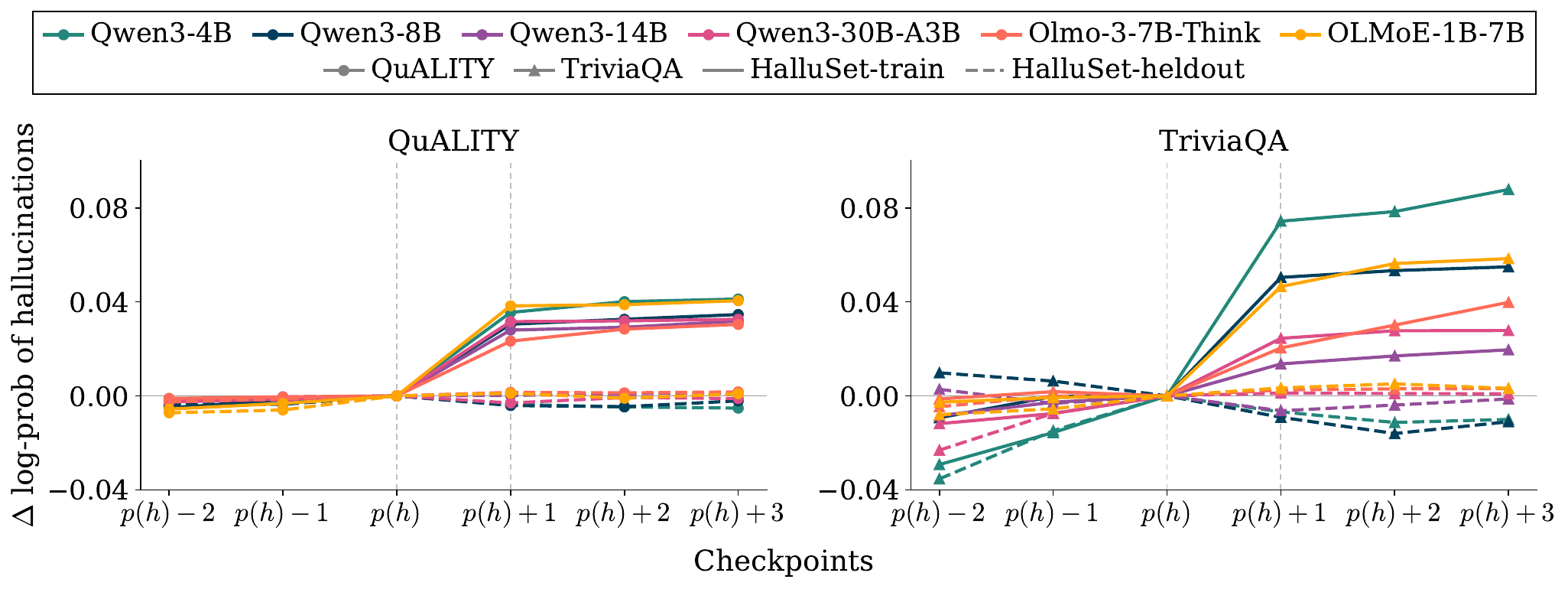} \vspace{-1.5mm}
\caption{Relative change in the log-probability of generating hallucinations from correct and held-out rollouts, compared to $M_{p(h)}$, the checkpoint immediately before the hallucination point. The likelihood of hallucinations from \texttt{HalluSet-train} increases sharply after the corresponding training update, while those from \texttt{HalluSet-heldout} show no comparable increase.}
\label{fig:logits}
\end{figure}

\begin{figure*}
    \centering
    \small
    \begin{minipage}[t]{0.47\textwidth}
        \captionsetup{type=table}
        \centering
        \renewcommand{\arraystretch}{0.9}
        \setlength{\tabcolsep}{6pt}
        \caption{Sentence-level hallucination rates, i.e., percentage of reasoning sentences that contain a hallucinated fact, for QuALITY and TriviaQA. Hallucinated facts are frequent across models, appearing in $25$--$55\%$ of reasoning sentences.}
        \label{tab:hallucination_rate_summary}
        \centering
        {
        \begin{tabular}{lcc}
        \toprule
        \textbf{Model} & \textbf{QuALITY} & \textbf{TriviaQA} \\
        \midrule
        Qwen3-4B        & 50.25 & 41.93 \\
        Qwen3-8B        & 50.84 & 32.21 \\
        Qwen3-14B       & 43.83 & 26.94 \\
        Qwen3-30B-A3B   & 48.54 & 28.53 \\
        Olmo-3-7B-Think & 53.73 & 38.90 \\
        OLMoE-1B-7B     & 54.88 & 46.18 \\
        \bottomrule
        \end{tabular}
        }
    \end{minipage} \hfill
    \begin{minipage}[t]{0.47\textwidth}
        \captionsetup{type=table}
        \centering
        \renewcommand{\arraystretch}{0.9}
        \setlength{\tabcolsep}{6pt}
        \caption{Memorization scores for backtracked and non-backtracked hallucinations on \texttt{HalluSet-train} for TriviaQA. Results show that backtracked (abbreviated as BT) hallucinations are memorized just as strongly as non-backtracked (non-BT) ones.}
        \label{tab:backtracking}
        {
        \begin{tabular}{lcc}
        \toprule
        \textbf{Model} & \textbf{Non-BT} & \textbf{BT} \\
        \midrule
        Qwen3-4B        & 3.46 & 3.28 \decp{0.18} \\
        Qwen3-8B        & 3.67 & 3.61 \decp{0.06} \\
        Qwen3-14B       & 3.77 & 3.63 \decp{0.14} \\
        Qwen3-30B-A3B   & 3.39 & 3.47 \incp{0.08} \\
        Olmo-3-7B-Think & 4.15 & 4.01 \decp{0.14} \\
        OLMoE-1B-7B     & 5.03 & 4.95 \decp{0.08} \\
        \bottomrule
        \end{tabular}
        }
    \end{minipage}
\end{figure*}

To further establish this memorization effect, we additionally conduct a non-MCQ-based evaluation. We track the log-probability of the hallucinations conditioned on the corresponding question for both \texttt{HalluSet-train} and \texttt{HalluSet-heldout}. \Cref{fig:logits} reports these for three checkpoints before and after the hallucination point; each value is reported as the difference from the log-probability at \(p(h)\). We find that \textbf{the probability of hallucinations in \texttt{HalluSet-train} increases substantially at the hallucination point, and this persists as training progresses.} Between $M_{p(h)}$ and $M_{p(h)+1}$, the log-probability increases sharply for \texttt{HalluSet-train}, while hallucinations in \texttt{HalluSet-heldout} show no comparable increase. This is consistent with the memorization trends observed in \Cref{tab:table1}.\footnote{Note that the values in \Cref{fig:logits} are not directly comparable to the memorization scores in \Cref{tab:table1}, as they are evaluated under two very different settings, albeit for the same question sets.}

\paragraph{LLMs memorize the hallucinated facts they backtrack from just as strongly as the others.} Often, reasoning chains propose false claims, but eventually backtrack from them (\Cref{fig:fig1}). In these cases, the reasoning chains are technically completely factual. But we hypothesize that backtracking is not sufficient, as LLMs will still memorize the false facts that they backtrack from.

Our results in \Cref{tab:backtracking} confirm our hypothesis. We report memorization scores for backtracked and non-backtracked hallucinations for TriviaQA.\footnote{We run this analysis only on TriviaQA, as the released checkpoints we use as initializations exhibit the frequent backtracking we want to study. In contrast, we cold-start QuALITY training using our own SFT data. We did not deliberately inject these ``suboptimal'' behaviors in this data to avoid biasing the analysis. As a result, QuALITY traces with backtracking are very limited for a meaningful comparison. See \Cref{fig:quality_tt_example} and \Cref{fig:trivia_tt_example} for representative reasoning traces from QuALITY and TriviaQA, respectively.} The two scores are nearly identical for all models, differing by at most $0.18$ points. This highlights that the reasoning behavior that makes RLVR effective on math and code carries none of its corrective value here. Instead, models memorize a hallucination just as strongly after retracting it as when they leave it uncorrected.

\paragraph{Memorization is strongest for hallucinations appearing in harder training questions.} We can estimate the difficulty of a question by the fraction of correct rollouts generated by the model for that question. For math, RL training is most beneficial for learning to solve difficult questions: it reinforces the strategies in the one or two correct rollouts that managed to generate the correct answer. For simple fact recall, on the other hand, there are no strategies to learn. 

\begin{figure}[t]
\centering
\includegraphics[width=1\linewidth]{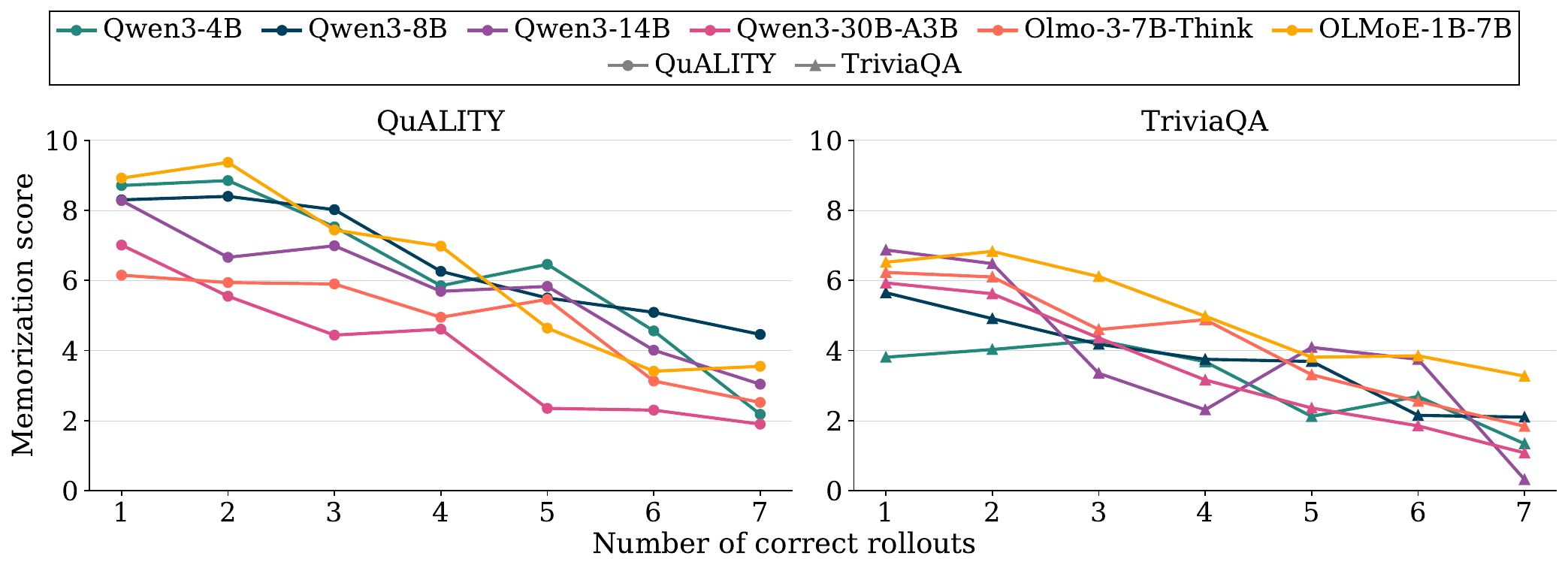}\vspace{-1mm}
\caption{Memorization score as a function of training-question difficulty, measured by the number of correct rollouts. Across all models, we find that hallucinations from harder questions, i.e., with fewer correct rollouts, generally have higher memorization scores.}
\label{fig:mem_vs_correct}
\end{figure}

In \Cref{fig:mem_vs_correct}, we plot the memorization scores for \texttt{HalluSet-train} across different numbers of correct rollouts that the originating question had during RL training. We find that the memorization scores increase with question difficulty. Therefore, contrary to math, the hardest training questions are the most detrimental, as they contribute the highest memorization of hallucinations.

A possible explanation for this trend comes from how GRPO computes advantages. Harder questions tend to have fewer correct rollouts. Since GRPO normalizes rewards within each group, when only a few rollouts are correct, these rollouts receive larger positive advantages than correct rollouts for easier questions. Hence, hallucinations in these rollouts may be reinforced more strongly, yielding higher memorization scores.

\vspace{-1mm}
\section{Ablations}
\label{sec:ablations}
\vspace{-1mm}

\paragraph{Do models become less likely to abstain after a hallucination is reinforced during training?} Next, we replace one of the distractor options in each MCQ with \textit{None of the above}, and evaluate whether models become less likely to choose this abstention option after a hallucination is reinforced during training. \Cref{tab:withNone_probs} reports the fraction of questions for which the model correctly selects this abstention option (higher is better) before and after $p(h)$, computed by averaging over checkpoints on either side of $p(h)$.

Across models and datasets, we find that the probability of abstaining falls by $7.23\%$ on average. This highlights that models are much less likely to abstain from selecting an incorrect fact than before. In fact, we see that the memorization scores for the hallucinated facts in this new setting are uniformly positive (\Cref{tab:withNone_memScores}), showing that these hallucinations are memorized strongly enough for the models to pick them over abstention. When we repeat this analysis on \texttt{HalluSet-heldout}, we find no such consistent trend in abstention after $p(h)$ (\Cref{tab:withNone_probs_heldout}).

\begin{table}[t]
\caption{Probability of selecting \textit{None of the above} on \texttt{HalluSet-train}, averaged over checkpoints before and after the hallucination point $p(h)$. Models become less likely to choose this abstention option after a self-hallucination is reinforced during training.}
\label{tab:withNone_probs}
\centering
\footnotesize
\setlength{\tabcolsep}{4pt}
\begin{tabular}{lcccccc}
\toprule
& \multicolumn{3}{c}{\textbf{QuALITY}}
& \multicolumn{3}{c}{\textbf{TriviaQA}} \\
\cmidrule(lr){2-4} \cmidrule(lr){5-7}
\textbf{Model}
& \multicolumn{1}{c}{Before $p(h) \uparrow$ }
& \multicolumn{1}{c}{After $p(h) \uparrow$}
& \multicolumn{1}{c}{Difference}
& \multicolumn{1}{c}{Before $p(h) \uparrow$}
& \multicolumn{1}{c}{After $p(h) \uparrow$}
& \multicolumn{1}{c}{Difference} \\
\midrule
Qwen3-4B        & 27.41 & 18.07 & \diffdown{9.34}  & 30.91 & 24.44 & \diffdown{6.47} \\
Qwen3-8B        & 34.84 & 30.45 & \diffdown{4.39}  & 32.51 & 27.10 & \diffdown{5.41} \\
Qwen3-14B       & 21.73 & 14.44 & \diffdown{7.29}  & 40.72 & 35.31 & \diffdown{5.41} \\
Qwen3-30B-A3B   & 41.59 & 23.95 & \diffdown{17.64} & 31.97 & 25.03 & \diffdown{6.94} \\
Olmo-3-7B-Think & 20.69 & 15.44 & \diffdown{5.25}  & 26.72 & 18.52 & \diffdown{8.20} \\
OLMoE-1B-7B     & 15.53 & 10.32 & \diffdown{5.21}  & 29.78 & 24.59 & \diffdown{5.19} \\
\bottomrule
\end{tabular}
\end{table}

\vspace{-2mm}
\paragraph{Do LLMs retain hallucinations memorized earlier in RL training as they memorize new ones?} We want to study if the recall of hallucinations in \texttt{HalluSet-train} persists as training progresses. To study this, we group hallucinations by the checkpoint \(M_{p(h)+1}\), i.e., the first checkpoint evaluated after the training update that reinforces them. We continue post-training the model for one additional epoch and, at the end of epoch 2, compare the recall of each hallucination group against its recall at the corresponding checkpoint \(M_j\).

\begin{figure}[t]
\centering
\includegraphics[width=1\linewidth]{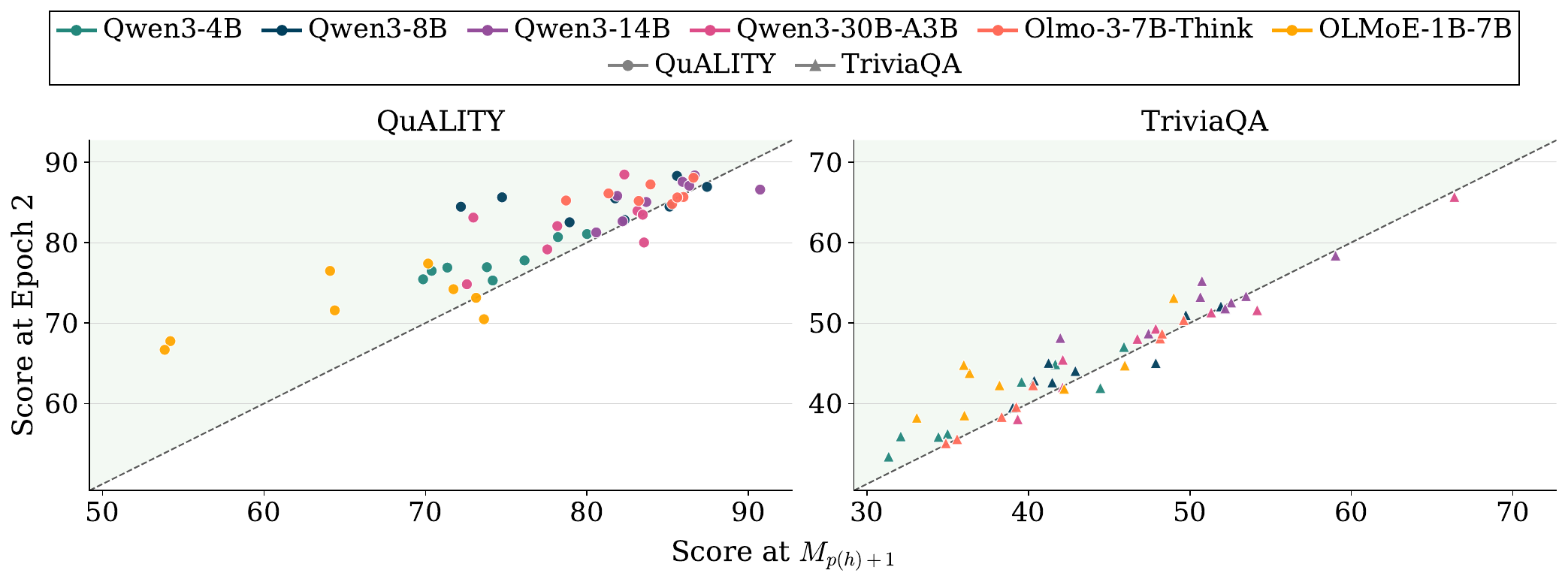}\vspace{-1mm}
\caption{Average scores on \texttt{HalluSet-train} at ${p(h)+1}$, the first evaluated checkpoint after the hallucination point, against average scores at the end of the second epoch. Each point denotes an evaluated checkpoint, so each model has multiple checkpoints shown. Points on or above the diagonal indicate that the score at the end of the second epoch is at or above the score at ${p(h)+1}$. Results show that models do not forget hallucinations memorized early during training. 
}
\label{fig:epoch2_scatter}
\end{figure}

\Cref{fig:epoch2_scatter} shows our results: points on or above the diagonal indicate checkpoints whose scores are maintained or higher at epoch 2 than at ${p(h)+1}$. We observe that once a hallucination has been reinforced during training, later checkpoints generally maintain their recall at roughly the same level, rather than declining as training continues. Models therefore retain hallucinations memorized early in RL while continuing to learn new ones as training progresses. See \Cref{tab:epoch2_scores} for detailed scores.

\begin{wraptable}{r}{0.4\linewidth}
\vspace{-14pt}
\centering
\small
\setlength{\tabcolsep}{4pt}
\caption{Memorization scores for QuALITY on \texttt{HalluSet-train} with and without a KL penalty. Memorization persists under KL regularization, with an average reduction of only $0.52$ points.}\vspace{-2mm}
\label{tab:kl_ablation}
\begin{tabular}{lcc}
\toprule
\textbf{Model} & \textbf{W/O KL} & \textbf{With KL} \\
\midrule
Qwen3-4B        & 5.85 & 5.13 \decp{0.72} \\
Qwen3-8B        & 6.14 & 5.34 \decp{0.80} \\
Qwen3-14B       & 5.52 & 5.01 \decp{0.51} \\
Qwen3-30B-A3B   & 4.44 & 4.58 \incp{0.14} \\
Olmo-3-7B-Think & 5.11 & 4.61 \decp{0.50} \\
OLMoE-1B-7B     & 6.88 & 6.15 \decp{0.73} \\
\bottomrule
\end{tabular}
\vspace{-8pt}
\end{wraptable}

\paragraph{Does memorization persist under a KL penalty?} While recent works implement GRPO without KL regularization \citep{yu2026dapo}, a KL penalty may suppress memorization by discouraging the updated policy from drifting too far from its initialization. To study this, we repeat our experiment on QuALITY\footnote{As noted in \S\ref{sec:questionGenerationDetails}, the pipeline costs approximately $\$35$ per \texttt{HalluSet} for TriviaQA. Hence, we conduct this experiment only for QuALITY.} with a KL coefficient of $0.001$, following commonly used GRPO settings \citep{guo2025deepseek,jin2025search}. \\
\Cref{tab:kl_ablation} reports results comparing the memorization scores with and without the KL penalty. We find that the memorization effect persists across models, with an average drop of only $0.52$ points relative to training without a KL penalty. This shows that adding the KL loss does not fix the memorization issue during training.

\section{Discussion And Implications}

Our work identifies and systematically analyzes a new mechanism through which models learn hallucinations during post-training. In this paper, we focus on understanding (i) whether the recall of self-generated hallucinations increases during training, (ii) when this memorization is strongest, including whether it persists for backtracked facts and how it varies with question difficulty, and (iii) whether this memorization naturally reverses, e.g., with further training or under a KL penalty. Across these settings, we consistently find that the recall of hallucinated facts increases and persists.

Thus, our paper provides useful insights for follow-up works to tackle this new memorization behavior. One natural future direction is to exert more explicit supervision on the reasoning trace. Some recent work has explored using process rewards and explicit factuality verification of reasoning trace sentences \citep{chen2025learning,li2026reasoning,yuan2026knowledge,ren2026knowrl}. However, operationalizing this at scale faces obstacles, particularly for long reasoning chains that exhibit ``cognitive'' behaviors \citep{gandhi2025cognitive}. First, this would require defining factuality at a finer-grained level, say sentence- or fact-level. However, design choices about decomposition and verifiability are not straightforward \citep{min2023factscore,song2024veriscore}. Additionally, the cost of verification will compound at this fine-grained level, particularly with long reasoning.

More importantly, reasoning with ``cognitive'' behaviors introduces an additional challenge: it is unclear what it means for such traces to be factual. In \Cref{fig:fig1} (middle), the trace as a whole is arguably factual because the model eventually retracts all the false facts it proposes. At a localized level, however, those retracted statements remain false. In fact, our results in \Cref{tab:backtracking} show that they are memorized as strongly as non-retracted ones. It is therefore unclear how a solution should deal with such globally factual but locally non-factual claims. Penalizing \textit{any} non-factual statement might suppress the very ``cognitive'' behaviors that make reasoning effective. 

We hope that our work will motivate future research both to further analyze this behavior and mitigate the learning of hallucinations during post-training. 

\vspace{-2mm}
\section{Related Work}

\paragraph{Test-time scaling and RLVR.} Scaling test-time compute through long intermediate \textit{thoughts} has become a predominant way to improve LLMs on challenging tasks \citep{wei2022chain,snell2024scaling,bi2024forest,brown2024large,chen2024more,zhao2025majority,hu2026pacore,yang2026towards}, and an effective alternative to scaling model parameters \citep{snell2024scaling,liu2025can}. It has produced strong results on math, code, and scientific problem solving \citep{jaech2024openai,guo2025deepseek,su2025dualformer,pan2025survey,lin2026controlling}, with long CoTs eliciting behaviors such as verification and backtracking \citep{gandhi2025cognitive}. Model providers also train specifically for these capabilities \citep{team2024qwq,team2025kimi,guo2025deepseek}, primarily through RLVR, whose outcome-based rewards are simple to implement \citep{shen2025satori,stojanovski2026reasoning}. However, such rewards might lead to unintended consequences, including linguistic incoherence and unfaithful reasoning \citep{chen2025reasoning,pan2025survey,yuan2026knowledge}, overthinking on easy problems \citep{luo2026o1,chen2024not}, degraded performance in certain domains \citep{yang2026towards}, and shorter or less detailed responses in the long-form generation domain \citep{chen2025learning}. We identify one more such failure mode.

\paragraph{Reasoning models and hallucinations.} \citet{kalai2025language} argue that outcome-based signals implicitly encourage hallucinations, since benchmarks reward guessing but not abstention. Reasoning models have also been shown to hallucinate more than their non-reasoning counterparts \citep{yao2025reasoning,chen2025learning}, an effect attributed to high-variance gradients, entropy-induced randomness, and susceptibility to spurious local optima \citep{li2026reasoning}. We identify a related problem: models memorize the hallucinations they are rewarded for. Even so, knowledge-intensive data is used for RL with outcome-based rewards, both in industry and academia \citep{liu2024deepseek,blakeman2025nemotron,olmo2026olmo3,ma2026improving}. In such setups, the reward signal typically does not evaluate reasoning quality \citep{hong2025glm,microsoft2025mai,guo2025deepseek}. While this might improve test-set accuracy, we show that it also carries an unintended consequence.

\paragraph{Mitigating hallucinations in post-training.} Post-training methods have been proposed to improve factuality \citep{tian2024fine,lin2024flame,zhang2024self}, largely through SFT and offline RL such as DPO \citep{rafailov2023direct}, showing that it is important to optimize factuality of self-generated responses during post-training (i.e., on-policy), rather than fine-tuning on unfamiliar knowledge (e.g., reference facts), which can itself increase hallucination \citep{kang2025unfamiliar,gekhman2024does,zhang2024self}. More recent work incorporates process rewards, i.e., explicitly verify the factuality of each reasoning sentence  \citep{chen2025learning,li2026reasoning,yuan2026knowledge,ren2026knowrl}, but rely largely on external verifiers like FactScore \citep{min2023factscore} and VeriScore \citep{song2024veriscore} or LLM judges \citep{zheng2023judging}, and are correspondingly limited by verifier bias, cost, and scalability, leaving effective mitigation an open challenge. 

\vspace{-3mm}
\section{Conclusion} 
\vspace{-2mm}
We show that reasoning-centric post-training causes LLMs to memorize self-generated hallucinations, with recall of hallucinations reinforced during training increasing by $4.83\%$ on average. This effect is strongest for hallucinations appearing in the hardest training questions. Moreover, models memorize hallucinations even when they subsequently backtrack from them and become less likely to abstain from hallucinations once they are reinforced. Our findings thus show that training recipes that are effective for math and code can carry an unintended cost when applied to knowledge tasks.

\section*{Acknowledgements} 
We would like to thank Oliver Li for helpful discussions and feedback. This project was partially supported by NSF grant IIS-2433072. We gratefully acknowledge the use of the research computing resources of the Empire AI Consortium, Inc, with support from Empire State Development of the State of New York, the Simons Foundation, and the Secunda Family Foundation. Anirudh Atmakuru was gratefully supported by the 2025-2026 Foundational AI Fellowship.

\subsection*{AI use statement}

In this work, we used generative AI tools as part of our experimental pipeline to detect hallucinations, generate synthetic data for SFT, and generate our evaluation datasets. Additionally, we used generative AI tools to edit portions of the manuscript for clarity, readability, and formatting. We did not use generative AI tools for other tasks requiring disclosure. We reviewed all AI-assisted outputs and take responsibility for the final content of this work, including all text, claims, and artifacts produced with the aid of generative AI.

\subsection*{Reproducibility statement}

We describe details of our experimental setup, datasets, models, training procedure, and evaluation in \S\ref{sec:sec2}. We provide additional implementation and evaluation details in the Appendix (\S\ref{sec:additionalDatasetDetails}, \S\ref{sec:additionalTrainingDetails}, \S\ref{sec:questionGenerationDetails}) along with the prompts we use in \S\ref{sec:prompts}. 

\bibliography{iclr2027_conference}
\bibliographystyle{iclr2027_conference}

\appendix
\section{Appendix}

\subsection{Additional Dataset Details}
\label{sec:additionalDatasetDetails}
\paragraph{QuALITY.} The QuALITY dataset \citep{pang2022QuALITY} is a reading-comprehension dataset consisting of 4609 questions, spanning 265 books and articles, totaling approximately 1.5M tokens. All questions are in multiple-choice format. We randomly split these into 3609 questions for training (used for both SFT and RL) and 1000 held out as a test set to evaluate the performance of our trained models. From the training set, we set aside roughly 5\% (175 questions) as a validation set for SFT and RL. We adopt a closed-book setting, prompting models to answer the question, given the book's title and author (\Cref{fig:sft_template_QuALITY}). 

\paragraph{TriviaQA.} We match the scale of QuALITY and randomly sample 4609 questions from TriviaQA \citep{joshi2017triviaqa}, and apply the same train/validation/test partition as done for QuALITY. All questions are short-form QA, each paired with a ground-truth answer and a set of aliases, with the final answer judged correct via a substring match against the alias list.

\subsection{Additional Training Details}
\label{sec:additionalTrainingDetails}

\subsubsection{QuALITY}
\paragraph{CPT details.}
Following the recipe in \citet{yang2024synthetic}, we perform CPT on each base model on the stories from QuALITY for 10M tokens, using a context length of 2048, a batch size of 16, and full-parameter training with Fully Sharded Data Parallelism (FSDP, \citet{zhao2023pytorch}). We apply a linear learning-rate warmup over the first 5\% of steps followed by a cosine decay, with a peak learning rate of $4\times10^{-5}$ for all models (except OLMoE-1B-7B, for which we use $5\times10^{-5}$), selected on a sweep over $\{5\times10^{-6},\,7.5\times10^{-6},\,1\times10^{-5},\,1.75\times10^{-5},\,2.5\times10^{-5},\,3.5\times10^{-5},\,4\times10^{-5},\,5\times10^{-5}\}$. To mitigate the forgetting of pretrained knowledge, we perform replay with a rate of 0.1 using 1B RedPajama tokens \citep{weber2024redpajama}.

\textbf{SFT details.} Following CPT, we perform a round of cold-start SFT to teach models to follow instructions and the desired output format. To build the training data for SFT, we prompt GPT-4.1-mini with each training question, its gold answer, and the full source story to generate a reasoning trace explaining the answer (prompt in \Cref{fig:prompt_sft_data}). On average, these reasoning traces are 72.07 words and 436.56 characters long. Each training example follows the same output format used downstream, with the reasoning enclosed within \texttt{<think></think>} tags and the final answer in \texttt{<answer></answer>} tags (\Cref{fig:sft_template_QuALITY}). Following \citet{yang2024synthetic}, we train for one epoch with a linear learning-rate warmup followed by cosine decay to 0, with a peak learning rate of $4\times10^{-5}$ for all models (except Qwen3-30B-A3B, for which we use $6\times10^{-5}$), a batch size of 16, and a maximum generation length of 1024 tokens. We selected the peak learning rate and batch size by sweeping learning rates in $\{7\times10^{-5},\,4\times10^{-5},\,1\times10^{-5},\,8\times10^{-6},\,5\times10^{-6},\,2.5\times10^{-6},\,1\times10^{-6}\}$ and batch sizes in $\{32, 16, 8, 4\}$.

\textbf{RL details.} We start from the cold-started SFT checkpoint and post-train using GRPO. Following recent practices \citep{yu2026dapo}, we perform full parameter training using GRPO without a KL penalty, using a learning rate of $1\times10^{-6}$. We use a maximum prompt length and maximum generation length of 512 tokens each, and a training batch size of 8. All our experiments use 8 checkpoints saved at equal intervals from the first epoch of RL. We use two checkpoints from the second epoch (corresponding to epochs 1.5 and 2) to test for forgetting of memorized knowledge (see \S\ref{sec:ablations}). For each training question, we sample 8 rollouts at temperature 0.7 using vLLM \citep{kwon2023efficient}. The reward checks the answer within the \texttt{<answer></answer>} tags against the gold QuALITY label. A rollout receives a reward of 1 when both the format and the final answer are correct, 0.2 when only the format is correct, and 0 otherwise. We uniformly use 2 NVIDIA B200 GPUs for all models, except for Qwen3-30B-A3B, for which we use 4 GPUs.

\paragraph{Validation results.} We first verify that CPT succeeds without damaging the model's general capabilities. During CPT, we save checkpoints at 5M and 10M training tokens and measure the model's ability to recall stories using ROUGE-2 \citep{lin2004rouge} by randomly sampling 150-token sequences from 150 stories and generating completions of the next 100 tokens at a temperature of 0.1. ROUGE-2 rises with CPT, with larger models generally reproducing the completions more faithfully (\Cref{tab:QuALITY_cpt}). To ensure that CPT does not cause catastrophic forgetting of pretraining knowledge, we evaluate each checkpoint on Measuring Massive Multitask Language Understanding (MMLU, \citet{hendrycks2020measuring}) using the \texttt{lm-evaluation-harness} \citep{gao2024language}, and observe no substantial degradation (\Cref{tab:QuALITY_cpt}). We also evaluate performance on the held-out test set of 1000 questions using few-shot examples, since these models are not yet instruction-tuned, and find that performance is largely preserved. \Cref{tab:QuALITY_cpt} reports these results.

\begin{table}[ht]
\caption{ROUGE-2, MMLU scores, and QuALITY test-set accuracy across different checkpoints during CPT with QuALITY.}
\label{tab:QuALITY_cpt}
\centering
\footnotesize
\begin{tabular*}{\linewidth}{@{\extracolsep{\fill}}lcccc}
\toprule
\textbf{Model} & \textbf{Checkpoint} & \textbf{ROUGE-2} & \textbf{MMLU} & \textbf{QuALITY testset} \\
\midrule
\multirow{3}{*}{Qwen3-4B}
  & Qwen3-4B-Base      & 0.0183 & 74.91 & 44.13 \\
  & CPT (5M)  & 0.1389 & 72.93 & 43.70 \\
  & CPT (10M) & \textbf{0.3601} & 72.03 & 44.12 \\
\midrule
\multirow{3}{*}{Qwen3-8B}
  & Qwen3-8B-Base      & 0.0213 & 78.85 & 49.00 \\
  & CPT (5M)  & 0.2581 & 76.51 & 47.80 \\
  & CPT (10M) & \textbf{0.4935} & 76.90 & 47.00 \\
\midrule
\multirow{3}{*}{Qwen3-14B}
  & Qwen3-14B-Base      & 0.0198 & 82.16 & 49.80 \\
  & CPT (5M)  & 0.2997 & 79.75 & 49.20 \\
  & CPT (10M) & \textbf{0.6541} & 78.17 & 49.30 \\
\midrule
\multirow{3}{*}{Qwen3-30B-A3B}
  & Qwen3-30B-A3B-Base      & 0.0256 & 82.85 & 46.20 \\
  & CPT (5M)  & 0.2764 & 81.68 & 47.00 \\
  & CPT (10M) & \textbf{0.5527} & 82.16 & 46.70 \\
\midrule
\multirow{3}{*}{Olmo-3-7B-Think}
  & Olmo-3-7B-Think-Base      & 0.0182 & 76.02 & 44.90 \\
  & CPT (5M)  & 0.3313 & 77.58 & 44.60 \\
  & CPT (10M) & \textbf{0.6606} & 78.17 & 45.00 \\
\midrule
\multirow{3}{*}{OLMoE-1B-7B}
  & OLMoE-1B-7B-Base      & 0.0218 & 54.36 & 29.40 \\
  & CPT (5M)  & 0.2884 & 50.42 & 30.60 \\
  & CPT (10M) & \textbf{0.5806} & 51.88 & 30.60 \\
\bottomrule
\end{tabular*}
\end{table}

We also evaluate the post-trained checkpoints on the held-out MCQs, sampling at a temperature of 0.6, top-p of 0.95, a repetition penalty of 1.1, and a maximum of 1024 new tokens (\Cref{tab:QuALITY_val}). An increasing trend in scores suggests successful training and that it is less likely that the selected models have been trained on the data from QuALITY (an average improvement of $\sim 8.93$ points from Epoch 0 to Epoch 2 across models).

\begin{table}[ht]
\caption{Accuracy of different post-trained checkpoints on the 1000-question test set of QuALITY.}
\label{tab:QuALITY_val}
\centering
\small
\resizebox{\linewidth}{!}{%
\begin{tabular}{lccccc}
\toprule
\textbf{Model} & \textbf{Epoch 0 (SFT/pre-RL)} & \textbf{Epoch 0.5} & \textbf{Epoch 1} & \textbf{Epoch 1.5} & \textbf{Epoch 2} \\
\midrule
Qwen3-4B  & 42.7 & 50.2 & 53.7 & 53.8 & 54.2 \\
Qwen3-8B  & 47.5 & 54.9 & 54.2 & 57.1 & 57.4 \\
Qwen3-14B & 51.7 & 58.4 & 59.2 & 59.4 & 59.3 \\
Qwen3-30B-A3B & 49.0 & 54.5 & 56.9 & 57.1 & 59.2 \\
Olmo-3-7B-Think & 45.9 & 49.1 & 50.6 & 51.9 & 52.0 \\
OLMoE-1B-7B & 38.1 & 40.3 & 42.2 & 42.8 & 46.4 \\
\bottomrule
\end{tabular}
}
\end{table}

\subsubsection{TriviaQA}
\paragraph{Training details.} We start directly from the post-trained checkpoints released by model providers\footnote{OLMoE-1B-7B is not a \textit{thinking} model, but it is post-trained and can follow instructions to generate answers in the required format.}, prompting each model to reason before producing its final answer (\Cref{fig:sft_template_trivia}). We follow the same GRPO setup used for QuALITY. However, since we start from a more verbose reasoning model, we set the maximum generation length to 1024. Correctness of each rollout is judged by substring match against the provided aliases. 

\paragraph{Validation results.} We evaluate five checkpoints using substring matching on the 1000 held-out TriviaQA test questions, sampling outputs with temperature 0.6, top-p 0.95, repetition penalty 1.1, and a maximum of 2048 new tokens. \Cref{tab:trivia_val} reports these results. We also note that these scores (\Cref{tab:trivia_val}) do not increase as consistently as those on QuALITY (\Cref{tab:QuALITY_val}), suggesting that the selected models might have seen much of this data already, and therefore do not require CPT (an average improvement of $\sim 3.22$ points from epoch 0 to epoch 2 across models).

\begin{table}[ht]
\caption{Accuracy of different checkpoints on the 1000-question test set of TriviaQA.}
\label{tab:trivia_val}
\centering
\small
\resizebox{\linewidth}{!}{%
\begin{tabular}{lccccc}
\toprule
\textbf{Model} & \textbf{Epoch 0 (pre-RL)} & \textbf{Epoch 0.5} & \textbf{Epoch 1} & \textbf{Epoch 1.5} & \textbf{Epoch 2} \\
\midrule
Qwen3-4B & 52.9 & 53.1 & 53.7 & 53.9 & 53.8 \\
Qwen3-8B & 65.7 & 64.8 & 65.6 & 67.7 & 67.9 \\
Qwen3-14B & 71.7 & 72.1 & 72.3 & 72.0 & 72.8 \\
Qwen3-30B-A3B & 73.1	& 73.7	&75.5	&78.4	&78.8 \\
Olmo-3-7B-Think & 53.6 & 54.9 & 57.3 & 58.7 & 58.9 \\
OLMoE-1B-7B & 53.7 & 55.3 & 56.3 & 57.1 & 57.8 \\
\bottomrule
\end{tabular}
}
\end{table}

\vspace{-2mm}
\subsection{Generating \texttt{HalluSet} Questions} 
\label{sec:questionGenerationDetails}
We break down the question generation process into two stages: 1) detecting hallucinations from the model's training rollout and 2) converting the hallucinations into MCQs. To detect hallucinations for models trained on QuALITY, we prompt GPT-4.1-mini with the training question, the gold answer, the model's reasoning, and the original story/article to identify a hallucinated fact (prompt in \Cref{fig:prompt_hall_detection_QuALITY}). We then convert this hallucination into an MCQ (prompt in \Cref{fig:prompt_question_gen_QuALITY}).

For TriviaQA, there is no reference text. We first prompt GPT-4.1-mini with the question, the model's reasoning, and the answer to extract three candidate hallucinations (prompt in \Cref{fig:prompt_hall_detection_trivia}). We then verify each with a web search, using the tool-use interface (\texttt{web\_search\_preview}) (prompt in \Cref{fig:prompt_web_check}). We select one verified false fact at random for question generation and discard those where all three candidates turn out to be correct. We then convert this hallucination into an MCQ (prompt in \Cref{fig:prompt_question_gen_trivia}). For the backtracking analysis in \S\ref{sec:results} (\Cref{tab:backtracking}), we repeat this sampling and question-generation procedure separately rather than partitioning the hallucinations used for analysis in \Cref{tab:table1}, to ensure that the number of evaluated questions remains roughly comparable across settings. For each rollout we independently extract three backtracked and three non-backtracked candidate hallucinations (prompt in \Cref{fig:prompt_hall_detection_trivia_backtrack}), verify them using the same web-search procedure, and sample one verified false fact from each category for MCQ generation. 

We adopt a two-step procedure for detecting hallucinations on TriviaQA since running web verification directly over long explanations generated by reasoning models is expensive. The complete process costs approximately $\$10$ for each set in QuALITY and approximately $\$35$ for TriviaQA. 

\paragraph{Selecting rollouts.} To construct \texttt{HalluSet-train}, we randomly choose a model-generated rollout with the correct final answer (if one exists), and apply the above procedure. We ignore questions for which the model generated no correct rollout. We also ignore questions for which all rollouts were correct, as this does not contribute to gradient updates. \texttt{HalluSet-heldout} serves as a control. We take the $1000$ held-out test questions from each dataset, split them into $8$ disjoint groups of $125$, and assign one group to each checkpoint $M_0,\dots, M_7$. We sample rollouts from the corresponding checkpoint for each group and identify a hallucination exactly the same way. Because these questions are held-out, no checkpoint is trained or rewarded on them.

\paragraph{Inference.} We evaluate each checkpoint on \texttt{HalluSet-train} and \texttt{HalluSet-heldout} using the same sampling parameters used for the test-set questions in \S\ref{sec:additionalTrainingDetails}.

\subsection{Additional Results}

\label{sec:additionalResults}
\paragraph{How often do models hallucinate in their reasoning traces?} The memorization effect we observe is less significant if models rarely hallucinate in their reasoning traces. To assess this, we randomly sample 200 correct rollouts from each model that were used to generate \texttt{HalluSet-train} and label every sentence in the reasoning trace as hallucinated or not (prompt in \Cref{fig:prompt_sentence_fact_check}). Across models, up to $55\%$ of the sentences produced in correct rollouts are hallucinated (\Cref{tab:hallucination_rate}). These constitute a substantial fraction of the reasoning traces, and since they also carry high memorization scores (\Cref{tab:table1}), this makes the effect we study broadly consequential.

\begin{table}[ht]
\caption{Hallucination rates in reasoning traces over 200 randomly sampled correct rollouts per model on QuALITY and TriviaQA.}

\label{tab:hallucination_rate}
\centering
\footnotesize
\setlength{\tabcolsep}{3pt}
\begin{tabular}{lccc@{\hskip 6pt}ccc}
\toprule
& \multicolumn{3}{c}{\textbf{QuALITY}}
& \multicolumn{3}{c}{\textbf{TriviaQA}} \\
\cmidrule(lr){2-4}\cmidrule(lr){5-7}
\textbf{Model}
& \makecell{\textbf{Sentences}\\\textbf{evaluated}}
& \makecell{\textbf{Hallucinated}\\\textbf{sentences}}
& \makecell{\textbf{Hallucination}\\\textbf{rate (\%)}}
& \makecell{\textbf{Sentences}\\\textbf{evaluated}}
& \makecell{\textbf{Hallucinated}\\\textbf{sentences}}
& \makecell{\textbf{Hallucination}\\\textbf{rate (\%)}} \\
\midrule
Qwen3-4B        & 408 & 205 & \textbf{50.25} & 2287 & 959 & \textbf{41.93} \\
Qwen3-8B        & 478 & 243 & \textbf{50.84} & 2620 & 844 & \textbf{32.21} \\
Qwen3-14B       & 575 & 252 & \textbf{43.83} & 2642 & 712 & \textbf{26.94} \\
Qwen3-30B-A3B   & 515 & 250 & \textbf{48.54} & 2411 & 688 & \textbf{28.53} \\
Olmo-3-7B-Think & 549 & 295 & \textbf{53.73} & 2514 & 978 & \textbf{38.90} \\
OLMoE-1B-7B     & 574 & 315 & \textbf{54.88} & 1791 & 827 & \textbf{46.18} \\
\bottomrule
\end{tabular}
\end{table}

\paragraph{Validity of distractor options in HalluSet.} We manually annotated 200 examples to verify that our models and prompts generated plausible distractors, and to verify whether the generated MCQs include the factually correct answers as distractors. For QuALITY, the ground truth answer did not appear in the distractors. This is because of two reasons: often the hallucination concerned a completely new event or property for which there did not exist a ground truth in the story. Furthermore, we used GPT-4.1-mini to generate distractors without providing it access to the story (\Cref{fig:prompt_question_gen_QuALITY}). For TriviaQA, we sampled 200 random MCQs from our HalluSet-train and annotated whether any of the distractor options were the ground truth. We found that the factual ground truth appeared in 39/200 questions (19.5\%). We deliberately did not enforce the exclusion of factually correct distractors, as their presence makes our results stronger: if the model increasingly selected the hallucination despite having the correct answer available, it provides stronger evidence of hallucination memorization. Among these 39 questions, the number of times the model selected the factual ground truth decreased from 27 initially to 15 at epoch 1, while the number of times it selected the hallucination over the ground truth increased from 8 to 14, respectively.

\paragraph{Additional results on memorization scores.}
\Cref{tab:memorization_scores_summary} reports the average recall before and after $p(h)$ underlying the memorization scores in \Cref{tab:table1}. The average difference on \textit{Before p(h)} between the train and held-out sets is $6.21$ points on QuALITY and $1.77$ points on TriviaQA, suggesting that the two sets have broadly comparable prior recall and difficulty. See full scores in \Cref{tab:per_bucket_scores} (\texttt{HalluSet-train}) and \Cref{tab:per_bucket_scores_neutral} (\texttt{HalluSet-heldout}). Additionally, for our experiment on backtracking in \Cref{tab:backtracking}, we conduct a paired permutation test, pairing the backtracked and non-backtracked hallucinations that come from the same rollout. We find no statistically significant difference between backtracked and non-backtracked hallucinations for any model (all $p>0.05$).

\begin{table}[ht]
\centering
\renewcommand{\arraystretch}{1.35}
\setlength{\tabcolsep}{7pt}
\small
\caption{Memorization results for \texttt{HalluSet-train} and \texttt{HalluSet-heldout} across models and datasets. We report the average recall immediately before the hallucination point $p(h)$, average recall after $p(h)$, and the resulting memorization score.}

\resizebox{\linewidth}{!}{%
\begin{tabular}{l|NNN|NNN}
\toprule
\textcolor{headertxt}{}
& \multicolumn{3}{c|}{\textcolor{headertxt}{\textbf{Train-time} (\texttt{\textbf{HalluSet-train}})}}
& \multicolumn{3}{c}{\textcolor{headertxt}{\textbf{Held-out} (\texttt{\textbf{HalluSet-heldout}})}} \\[-2pt]

\textcolor{headertxt}{\textbf{Model}}
& \textcolor{headertxt}{\textbf{Before $p(h)$}}
& \textcolor{headertxt}{\textbf{After $p(h)$}}
& \textcolor{headertxt}{\textbf{mem. score}}
& \textcolor{headertxt}{\textbf{Before $p(h)$}}
& \textcolor{headertxt}{\textbf{After $p(h)$}}
& \textcolor{headertxt}{\textbf{mem. score}} \\
\midrule

\rowcolor{headergray}
\multicolumn{7}{c}{\textcolor{headertxt}{\textbf{QuALITY}}} \\

Qwen3-4B
& 69.83 & 75.68 & 5.85
& 77.97 & 78.86 & 0.89 \\

Qwen3-8B
& 76.09 & 82.23 & 6.14
& 82.98 & 83.22 & 0.24 \\

Qwen3-14B
& 79.28 & 84.80 & 5.52
& 84.30 & 84.88 & 0.58 \\

Qwen3-30B-A3B
& 74.94 & 79.38 & 4.44
& 82.00 & 82.56 & 0.56 \\

Olmo-3-7B-Think
& 80.03 & 85.14 & 5.11
& 84.70 & 84.68 & -0.02 \\

OLMoE-1B-7B
& 61.11 & 67.99 & 6.88
& 66.60 & 67.15 & 0.55 \\

\midrule

\rowcolor{headergray}
\multicolumn{7}{c}{\textcolor{headertxt}{\textbf{TriviaQA}}} \\

Qwen3-4B
& 35.12 & 38.31 & 3.19
& 39.84 & 39.14 & -0.70 \\

Qwen3-8B
& 41.15 & 44.86 & 3.71
& 46.37 & 45.61 & -0.76 \\

Qwen3-14B
& 48.45 & 52.01 & 3.56
& 46.88 & 46.97 & 0.09 \\

Qwen3-30B-A3B
& 44.84 & 48.61 & 3.77
& 45.47 & 45.24 & -0.23 \\

Olmo-3-7B-Think
& 38.10 & 42.34 & 4.24
& 41.42 & 42.08 & 0.66 \\

OLMoE-1B-7B
& 34.54 & 40.12 & 5.58
& 32.86 & 33.00 & 0.14 \\

\bottomrule
\end{tabular}%
}
\label{tab:memorization_scores_summary}
\end{table}

\paragraph{Additional results on \textit{None of the Above} ablation.} \Cref{tab:withNone_memScores} shows the memorization scores on \texttt{HalluSet-train} with one distractor replaced by \textit{None of the Above} as detailed in \S\ref{sec:ablations}. We see that the memorization scores are uniformly positive, showing that these hallucinations are memorized strongly enough for the models to pick them over abstention. Additionally, for comparison, \Cref{tab:withNone_probs_heldout} reports the abstention analysis on \texttt{HalluSet-heldout}. Unlike \texttt{HalluSet-train} in \Cref{tab:withNone_probs}, there is no consistent trend in selecting \textit{None of the above} after $p(h)$ compared to before, with an average difference of only $0.07$ percentage points across models and datasets.

\begin{table}[ht]
\caption{Memorization scores on \texttt{HalluSet-train} when a distractor is replaced with \textit{None of the above}. Memorization scores are positive even when models are given the option to abstain.}
\label{tab:withNone_memScores}
\centering
\small
\setlength{\tabcolsep}{6pt}
\begin{tabular}{lcc}
\toprule
\textbf{Model} & \textbf{QuALITY} & \textbf{TriviaQA} \\
\midrule
Qwen3-4B        & 3.49 & 3.01 \\
Qwen3-8B        & 3.53 & 2.18 \\
Qwen3-14B       & 4.11 & 3.49 \\
Qwen3-30B-A3B   & 3.97 & 3.07 \\
Olmo-3-7B-Think & 4.17 & 4.34 \\
OLMoE-1B-7B     & 4.25 & 4.28 \\
\bottomrule
\end{tabular}
\end{table}

\begin{table}[ht]
\caption{Probability of selecting \textit{None of the above} on \texttt{HalluSet-heldout}, averaged over checkpoints before and after the hallucination point $p(h)$. There is no consistent change in abstention after $p(h)$ compared to before.} 

\label{tab:withNone_probs_heldout}
\centering
\footnotesize
\setlength{\tabcolsep}{6pt}
\begin{tabular}{lcccccc}
\toprule
& \multicolumn{3}{c}{\textbf{QuALITY}}
& \multicolumn{3}{c}{\textbf{TriviaQA}} \\
\cmidrule(lr){2-4} \cmidrule(lr){5-7}
\textbf{Model}
& \multicolumn{1}{c}{Before $p(h)$}
& \multicolumn{1}{c}{After $p(h)$}
& \multicolumn{1}{c}{Difference}
& \multicolumn{1}{c}{Before $p(h)$}
& \multicolumn{1}{c}{After $p(h)$}
& \multicolumn{1}{c}{Difference} \\
\midrule
Qwen3-4B        & 22.05 & 19.01 &  3.04 & 26.09 & 24.23 &  1.86 \\
Qwen3-8B        & 31.56 & 33.68 & -2.12 & 30.69 & 31.34 & -0.65 \\
Qwen3-14B       & 24.12 & 20.98 &  3.14 & 32.46 & 29.91 &  2.55 \\
Qwen3-30B-A3B   & 33.83 & 38.81 & -4.98 & 33.05 & 30.67 &  2.38 \\
Olmo-3-7B-Think & 15.29 & 17.03 & -1.74 & 22.86 & 19.39 &  3.47 \\
OLMoE-1B-7B     &  8.87 & 12.67 & -3.80 & 24.71 & 27.04 & -2.33 \\
\bottomrule
\end{tabular}
\end{table}

\paragraph{Ablation with hallucinations from incorrect rollouts.} As an additional control, we repeat our analysis using hallucinations sampled from rollouts with incorrect final answers, denoted \texttt{HalluSet-neg}. These hallucinations generally have memorization scores close to or less than zero (\Cref{tab:incorrect_mem}), and models become more likely to select \textit{None of the above} after $p(h)$ than before (\Cref{tab:withNone_neg}); both trends are opposite to those observed for \texttt{HalluSet-train}.

\begin{table}[ht]
\caption{Memorization scores on \texttt{HalluSet-neg}. Hallucinations carried by an incorrect final answer yield negative memorization scores in general.}
\label{tab:incorrect_mem}
\centering
\small
\setlength{\tabcolsep}{6pt}
\begin{tabular}{lcc}
\toprule
\textbf{Model} & \textbf{QuALITY} & \textbf{TriviaQA} \\
\midrule
Qwen3-4B        & 0.61  & -3.01 \\
Qwen3-8B        & 1.15  & -2.18 \\
Qwen3-14B       & -0.40 & -2.38 \\
Qwen3-30B-A3B   & -1.30 & -2.55 \\
Olmo-3-7B-Think & -0.23 & -0.09 \\
OLMoE-1B-7B     & -0.86 & -1.69 \\
\bottomrule
\end{tabular}
\end{table}

\begin{table}[ht]
\caption{Probability of selecting \textit{None of the above} on \texttt{HalluSet-neg}, averaged over checkpoints before and after the hallucination point $p(h)$.}
\label{tab:withNone_neg}
\centering
\footnotesize
\setlength{\tabcolsep}{6pt}
\begin{tabular}{lcccccc}
\toprule
& \multicolumn{3}{c}{\textbf{QuALITY}}
& \multicolumn{3}{c}{\textbf{TriviaQA}} \\
\cmidrule(lr){2-4} \cmidrule(lr){5-7}
\textbf{Model}
& \multicolumn{1}{c}{Before $p(h)$}
& \multicolumn{1}{c}{After $p(h)$}
& \multicolumn{1}{c}{Difference}
& \multicolumn{1}{c}{Before $p(h)$}
& \multicolumn{1}{c}{After $p(h)$}
& \multicolumn{1}{c}{Difference} \\
\midrule
Qwen3-4B        & 20.19 & 31.24 & -11.05 & 24.13 & 28.45 & -4.32 \\
Qwen3-8B        & 29.39 & 37.61 & -8.22  & 27.55 & 32.36 & -4.81 \\
Qwen3-14B       & 14.80 & 26.79 & -11.99 & 34.05 & 39.13 & -5.08 \\
Qwen3-30B-A3B   & 25.98 & 46.63 & -20.65 & 31.33 & 36.48 & -5.15 \\
Olmo-3-7B-Think & 16.85 & 22.70 & -5.85  & 18.86 & 27.66 & -8.80 \\
OLMoE-1B-7B     & 10.89 & 15.33 & -4.44  & 24.31 & 29.24 & -4.93 \\
\bottomrule
\end{tabular}
\end{table}

\paragraph{Sensitivity to MCQ option ordering and distractors.} To test whether our results are sensitive to the MCQ format, we randomly shuffle the options of all MCQs in \texttt{HalluSet-train} and repeat inference across all models and datasets, keeping the questions and distractors unchanged. \Cref{tab:shuffled_mem} reports the resulting memorization scores. Compared to \Cref{tab:table1}, the scores remain very similar, with a mean absolute change of 0.13 and a maximum change of 0.23. We also regenerate the distractor options for all MCQs in \texttt{HalluSet-train} for Qwen3-4B and Olmo-3-7B-Think, keeping the questions unchanged. On QuALITY, the memorization scores change from 5.85 to 5.72 for Qwen3-4B and from 5.11 to 5.30 for Olmo-3-7B-Think; on TriviaQA, they change from 3.19 to 3.27 and from 4.24 to 4.03, respectively (mean absolute change of 0.15). Conceptually, our memorization score's design is, in fact, robust to such potential confounders. This is because it does not report absolute scores for a specific checkpoint but rather the change in this score for different checkpoints throughout training. Together, this shows that our results are robust to variations in distractor construction and answer-option ordering.

\begin{table}[ht] 
\caption{Memorization scores on \texttt{HalluSet-train} after randomly shuffling the MCQ options. Scores remain close to those in \Cref{tab:table1}, with a mean absolute change of 0.13.}
\label{tab:shuffled_mem}
\centering
\small
\setlength{\tabcolsep}{6pt}
\begin{tabular}{lcc}
\toprule
\textbf{Model} & \textbf{QuALITY} & \textbf{TriviaQA} \\
\midrule
Qwen3-4B        & 5.77 & 3.07 \\
Qwen3-8B        & 6.25 & 3.88 \\
Qwen3-14B       & 5.29 & 3.49 \\
Qwen3-30B-A3B   & 4.60 & 3.69 \\
Olmo-3-7B-Think & 5.20 & 4.32 \\
OLMoE-1B-7B     & 6.72 & 5.81 \\
\bottomrule
\end{tabular}
\end{table}

\paragraph{Memorization beyond GRPO.} We repeat our main experiments using PPO \citep{schulman2017proximal}, keeping the rest of the experimental setup unchanged. On \texttt{HalluSet-train}, the resulting memorization scores with PPO (\Cref{tab:ppo_mem}) are close to those obtained with GRPO in \Cref{tab:table1}. Conceptually, this is expected because both algorithms use outcome-based rewards, and the trajectories with hallucinations may be assigned positive rewards as long as the final outcome is correct. So, even though PPO and GRPO use different ways of estimating the advantage and objectives, they optimize the expected value of the same flawed reward.

\begin{table}[ht]
\caption{Memorization scores on \texttt{HalluSet-train} with PPO. Scores remain close to those obtained with GRPO in \Cref{tab:table1}.}
\label{tab:ppo_mem}
\centering
\small
\setlength{\tabcolsep}{6pt}
\begin{tabular}{lcc}
\toprule
\textbf{Model} & \textbf{QuALITY} & \textbf{TriviaQA} \\
\midrule
Qwen3-4B        & 5.90 & 3.56 \\
Qwen3-8B        & 5.92 & 4.02 \\
Qwen3-14B       & 5.81 & 3.87 \\
Qwen3-30B-A3B   & 4.61 & 3.61 \\
Olmo-3-7B-Think & 4.80 & 3.82 \\
OLMoE-1B-7B     & 6.35 & 6.01 \\
\bottomrule
\end{tabular}
\end{table}

\paragraph{Reliability of LLM-based annotation.} Frontier LLMs have been widely used for factuality evaluation \citep{min2023factscore,zheng2023judging,wei2024long} and for generating MCQ-style questions \citep{wang2024mmlu}. Following these works, we use frontier LLMs to detect hallucinations (given a reference text or web search) and convert them into MCQs, refining all prompts through several rounds of prompt engineering and manual spot-checking, and providing few-shot examples. Frontier LLMs have also been shown to reliably answer questions and verify claims against a given document \citep{shaham2023zeroscrolls,jacovi2025facts} and, with web access, to verify open-domain facts \citep{wei2024long,wei2025browsecomp} closely matching the settings in which we detect hallucinations. Additionally, one of the authors manually verified 200 randomly sampled TriviaQA facts labeled as hallucinated against the web and found a precision of 98\%. Similarly, one of the authors also manually verified the backtracking labels for 200 randomly sampled claims from TriviaQA, finding an annotation accuracy of 87\% for backtracked claims and 91\% for non-backtracked claims. These results provide further evidence that frontier LLMs can reliably perform these largely non-subjective tasks when given clear instructions.

\clearpage

\begingroup
\renewcommand{\arraystretch}{0.45}
\setlength{\LTcapwidth}{\textwidth}
\captionsetup{font=normalsize}  
\footnotesize
\setlength{\tabcolsep}{3pt}
\begin{longtable}{cccccccccc}
\caption{Scores on \texttt{HalluSet-train}, binned by hallucination point. Rows denote checkpoints. $S_t$ denotes questions whose corresponding hallucinations were generated between checkpoints $M_{t-1}$ and $M_t$ (counts in parentheses). Bold cells denote the first evaluated checkpoint after the hallucination point for each column's facts. Scores generally increase after the corresponding hallucinations are generated during training.}

\label{tab:per_bucket_scores}\\
\toprule
\endfirsthead
\multicolumn{10}{c}{\tablename~\thetable{} -- \textit{continued from previous page}}\\
\toprule
\endhead
\midrule
\multicolumn{10}{r}{\textit{continued on next page}}\\
\endfoot
\bottomrule
\endlastfoot
\multicolumn{10}{c}{\textbf{QuALITY -- Qwen3-4B}} \\*
\midrule
\textbf{Eval ckpt} & $S_1$ (335) & $S_2$ (314) & $S_3$ (267) & $S_4$ (229) & $S_5$ (222) & $S_6$ (211) & $S_7$ (192) & $S_8$ (185) & Overall (1955) \\*
\midrule
\setrange{65.17}{77.80}%
$M_0$ & 66.27 & 62.42 & 64.42 & 62.88 & 68.92 & 63.98 & 68.23 & 65.41 & \heat{65.17} \\
$M_1$ & \seen{\textbf{69.85}} & 70.38 & 66.29 & 64.63 & 69.37 & 71.56 & 67.19 & 72.97 & \heat{69.00} \\
$M_2$ & \seen{70.15} & \seen{\textbf{70.38}} & 69.66 & 67.25 & 71.17 & 68.25 & 71.88 & 69.73 & \heat{69.82} \\
$M_3$ & \seen{77.31} & \seen{75.80} & \seen{\textbf{74.16}} & 65.94 & 74.77 & 73.93 & 68.75 & 78.92 & \heat{73.96} \\
$M_4$ & \seen{74.03} & \seen{76.11} & \seen{75.66} & \seen{\textbf{73.80}} & 71.62 & 73.93 & 72.92 & 76.22 & \heat{74.37} \\
$M_5$ & \seen{75.22} & \seen{79.30} & \seen{76.03} & \seen{72.05} & \seen{\textbf{76.13}} & 71.88 & 69.79 & 78.38 & \heat{75.12} \\
$M_6$ & \seen{77.01} & \seen{77.71} & \seen{73.41} & \seen{69.00} & \seen{72.07} & \seen{\textbf{78.20}} & 69.52 & 78.38 & \heat{74.65} \\
$M_7$ & \seen{78.51} & \seen{77.71} & \seen{73.78} & \seen{71.62} & \seen{76.13} & \seen{77.25} & \seen{\textbf{71.35}} & 76.22 & \heat{75.60} \\
$M_8$ & \seen{78.21} & \seen{81.21} & \seen{76.03} & \seen{73.36} & \seen{78.38} & \seen{77.73} & \seen{76.56} & \seen{\textbf{80.00}} & \heat{77.80} \\
\midrule
\multicolumn{10}{c}{\textbf{QuALITY -- Qwen3-8B}} \\*
\midrule
\textbf{Eval ckpt} & $S_1$ (302) & $S_2$ (285) & $S_3$ (241) & $S_4$ (223) & $S_5$ (191) & $S_6$ (198) & $S_7$ (162) & $S_8$ (148) & Overall (1750) \\*
\midrule
\setrange{70.51}{83.31}%
$M_0$ & 70.53 & 67.02 & 75.10 & 73.09 & 76.96 & 65.15 & 72.22 & 62.84 & \heat{70.51} \\
$M_1$ & \seen{\textbf{72.19}} & 74.39 & 75.93 & 74.89 & 73.30 & 72.73 & 74.69 & 68.92 & \heat{73.54} \\
$M_2$ & \seen{73.51} & \seen{\textbf{74.74}} & 78.01 & 78.03 & 75.92 & 73.74 & 77.16 & 75.68 & \heat{75.71} \\
$M_3$ & \seen{75.83} & \seen{74.39} & \seen{\textbf{81.74}} & 76.68 & 80.10 & 75.25 & 81.48 & 77.70 & \heat{77.60} \\
$M_4$ & \seen{80.46} & \seen{78.60} & \seen{80.91} & \seen{\textbf{78.92}} & 80.10 & 75.25 & 83.33 & 81.76 & \heat{79.77} \\
$M_5$ & \seen{79.47} & \seen{81.05} & \seen{83.40} & \seen{79.82} & \seen{\textbf{87.43}} & 74.75 & 81.48 & 78.38 & \heat{80.74} \\
$M_6$ & \seen{85.76} & \seen{85.26} & \seen{85.06} & \seen{83.41} & \seen{85.34} & \seen{\textbf{82.32}} & 80.57 & 79.73 & \heat{83.86} \\
$M_7$ & \seen{82.12} & \seen{83.16} & \seen{82.16} & \seen{83.41} & \seen{86.91} & \seen{82.32} & \seen{\textbf{85.57}} & 78.41 & \heat{83.01} \\
$M_8$ & \seen{82.45} & \seen{84.21} & \seen{85.06} & \seen{82.51} & \seen{84.82} & \seen{78.79} & \seen{85.80} & \seen{\textbf{85.11}} & \heat{83.48} \\
\midrule
\multicolumn{10}{c}{\textbf{QuALITY -- Qwen3-14B}} \\*
\midrule
\textbf{Eval ckpt} & $S_1$ (277) & $S_2$ (236) & $S_3$ (183) & $S_4$ (180) & $S_5$ (147) & $S_6$ (139) & $S_7$ (133) & $S_8$ (112) & Overall (1407) \\*
\midrule
\setrange{78.08}{85.75}%
$M_0$ & 79.31 & 73.66 & 74.32 & 80.00 & 82.31 & 77.70 & 82.71 & 76.79 & \heat{78.08} \\
$M_1$ & \seen{\textbf{85.92}} & 75.69 & 77.42 & 77.78 & 80.95 & 79.86 & 81.95 & 75.89 & \heat{79.77} \\
$M_2$ & \seen{85.56} & \seen{\textbf{82.20}} & 78.42 & 80.56 & 82.31 & 83.45 & 83.46 & 83.04 & \heat{82.48} \\
$M_3$ & \seen{85.20} & \seen{78.39} & \seen{\textbf{81.87}} & 79.11 & 82.31 & 82.01 & 82.71 & 81.25 & \heat{81.68} \\
$M_4$ & \seen{89.89} & \seen{81.78} & \seen{81.33} & \seen{\textbf{86.67}} & 80.95 & 83.33 & 87.22 & 76.79 & \heat{84.13} \\
$M_5$ & \seen{88.09} & \seen{81.36} & \seen{83.61} & \seen{85.00} & \seen{\textbf{83.67}} & 82.05 & 87.97 & 78.57 & \heat{84.16} \\
$M_6$ & \seen{87.73} & \seen{77.97} & \seen{85.79} & \seen{85.00} & \seen{83.95} & \seen{\textbf{86.33}} & 86.22 & 77.68 & \heat{84.02} \\
$M_7$ & \seen{84.95} & \seen{79.66} & \seen{82.97} & \seen{85.00} & \seen{84.63} & \seen{86.33} & \seen{\textbf{90.72}} & 74.11 & \heat{84.60} \\
$M_8$ & \seen{85.39} & \seen{82.93} & \seen{85.42} & \seen{88.11} & \seen{85.31} & \seen{86.17} & \seen{93.22} & \seen{\textbf{80.57}} & \heat{85.75} \\
\midrule
\multicolumn{10}{c}{\textbf{QuALITY -- Qwen3-30B-A3B}} \\*
\midrule
\textbf{Eval ckpt} & $S_1$ (272) & $S_2$ (266) & $S_3$ (206) & $S_4$ (187) & $S_5$ (170) & $S_6$ (166) & $S_7$ (147) & $S_8$ (139) & Overall (1553) \\*
\midrule
\setrange{69.77}{79.52}%
$M_0$ & 70.06 & 68.42 & 70.87 & 67.91 & 75.88 & 68.67 & 71.43 & 64.75 & \heat{69.77} \\
$M_1$ & \seen{\textbf{72.96}} & 69.17 & 75.73 & 70.59 & 81.18 & 74.10 & 74.15 & 71.22 & \heat{73.37} \\
$M_2$ & \seen{81.25} & \seen{\textbf{72.56}} & 75.73 & 74.33 & 77.65 & 77.71 & 74.83 & 70.50 & \heat{75.85} \\
$M_3$ & \seen{76.10} & \seen{74.44} & \seen{\textbf{78.16}} & 74.87 & 77.00 & 78.92 & 80.27 & 82.01 & \heat{77.27} \\
$M_4$ & \seen{76.84} & \seen{73.31} & \seen{83.01} & \seen{\textbf{77.54}} & 77.76 & 79.52 & 80.27 & 78.42 & \heat{78.00} \\
$M_5$ & \seen{77.21} & \seen{72.18} & \seen{80.10} & \seen{77.01} & \seen{\textbf{83.53}} & 80.13 & 82.99 & 74.82 & \heat{78.05} \\
$M_6$ & \seen{77.21} & \seen{71.80} & \seen{79.61} & \seen{78.61} & \seen{84.12} & \seen{\textbf{83.12}} & 81.63 & 77.70 & \heat{78.62} \\
$M_7$ & \seen{77.21} & \seen{74.44} & \seen{80.10} & \seen{78.61} & \seen{81.18} & \seen{83.12} & \seen{\textbf{82.31}} & 77.70 & \heat{78.88} \\
$M_8$ & \seen{77.21} & \seen{73.68} & \seen{80.10} & \seen{78.61} & \seen{80.59} & \seen{83.12} & \seen{85.71} & \seen{\textbf{83.45}} & \heat{79.52} \\
\midrule
\multicolumn{10}{c}{\textbf{QuALITY -- Olmo-3-7B-Think}} \\*
\midrule
\textbf{Eval ckpt} & $S_1$ (338) & $S_2$ (316) & $S_3$ (305) & $S_4$ (276) & $S_5$ (256) & $S_6$ (258) & $S_7$ (237) & $S_8$ (201) & Overall (2187) \\*
\midrule
\setrange{76.73}{87.70}%
$M_0$ & 76.04 & 75.00 & 81.97 & 77.17 & 77.34 & 74.42 & 74.26 & 77.11 & \heat{76.73} \\
$M_1$ & \seen{\textbf{78.70}} & 77.22 & 80.97 & 81.16 & 77.73 & 75.58 & 83.12 & 77.61 & \heat{79.02} \\
$M_2$ & \seen{82.25} & \seen{\textbf{81.33}} & 81.61 & 82.97 & 82.42 & 74.42 & 77.64 & 75.62 & \heat{80.10} \\
$M_3$ & \seen{83.14} & \seen{82.91} & \seen{\textbf{83.93}} & 80.43 & 80.86 & 78.29 & 82.28 & 77.11 & \heat{81.39} \\
$M_4$ & \seen{83.73} & \seen{79.11} & \seen{82.30} & \seen{\textbf{86.59}} & 82.20 & 77.13 & 82.28 & 77.61 & \heat{81.55} \\
$M_5$ & \seen{86.98} & \seen{80.70} & \seen{85.90} & \seen{84.78} & \seen{\textbf{83.20}} & 80.62 & 83.12 & 76.62 & \heat{83.08} \\
$M_6$ & \seen{86.98} & \seen{84.81} & \seen{88.52} & \seen{86.59} & \seen{84.38} & \seen{\textbf{85.27}} & 83.08 & 81.09 & \heat{85.36} \\
$M_7$ & \seen{85.50} & \seen{84.81} & \seen{88.52} & \seen{85.87} & \seen{83.59} & \seen{83.33} & \seen{\textbf{85.97}} & 81.09 & \heat{85.03} \\
$M_8$ & \seen{90.24} & \seen{87.34} & \seen{89.51} & \seen{90.94} & \seen{86.33} & \seen{84.11} & \seen{85.65} & \seen{\textbf{85.59}} & \heat{87.70} \\
\midrule
\multicolumn{10}{c}{\textbf{QuALITY -- OLMoE-1B-7B}} \\*
\midrule
\textbf{Eval ckpt} & $S_1$ (345) & $S_2$ (351) & $S_3$ (323) & $S_4$ (306) & $S_5$ (305) & $S_6$ (283) & $S_7$ (254) & $S_8$ (268) & Overall (2435) \\*
\midrule
\setrange{55.11}{70.83}%
$M_0$ & 53.04 & 51.00 & 57.59 & 53.92 & 56.07 & 57.60 & 56.30 & 56.72 & \heat{55.11} \\
$M_1$ & \seen{\textbf{54.20}} & 54.42 & 59.75 & 58.82 & 55.41 & 61.84 & 61.81 & 53.36 & \heat{57.29} \\
$M_2$ & \seen{55.94} & \seen{\textbf{53.85}} & 59.13 & 62.42 & 63.28 & 65.37 & 65.75 & 55.97 & \heat{59.92} \\
$M_3$ & \seen{58.26} & \seen{61.82} & \seen{\textbf{64.09}} & 60.78 & 61.64 & 63.96 & 64.17 & 65.67 & \heat{62.38} \\
$M_4$ & \seen{62.90} & \seen{56.98} & \seen{64.40} & \seen{\textbf{64.38}} & 63.93 & 68.55 & 62.60 & 63.06 & \heat{63.20} \\
$M_5$ & \seen{63.77} & \seen{64.10} & \seen{68.11} & \seen{67.97} & \seen{\textbf{70.16}} & 69.61 & 63.78 & 66.79 & \heat{66.74} \\
$M_6$ & \seen{65.22} & \seen{61.25} & \seen{72.76} & \seen{68.30} & \seen{70.49} & \seen{\textbf{71.73}} & 63.39 & 68.66 & \heat{67.64} \\
$M_7$ & \seen{68.70} & \seen{66.10} & \seen{71.21} & \seen{72.22} & \seen{74.10} & \seen{71.73} & \seen{\textbf{73.62}} & 68.66 & \heat{70.64} \\
$M_8$ & \seen{64.35} & \seen{64.39} & \seen{69.35} & \seen{67.97} & \seen{72.46} & \seen{73.50} & \seen{67.32} & \seen{\textbf{73.13}} & \heat{68.83} \\
\midrule
\multicolumn{10}{c}{\textbf{TriviaQA -- Qwen3-4B}} \\*
\midrule
\textbf{Eval ckpt} & $S_1$ (98) & $S_2$ (93) & $S_3$ (96) & $S_4$ (84) & $S_5$ (92) & $S_6$ (80) & $S_7$ (81) & $S_8$ (92) & Overall (716) \\*
\midrule
\setrange{36.47}{39.15}%
$M_0$ & 43.00 & 43.01 & 37.50 & 34.52 & 33.70 & 31.25 & 32.10 & 34.78 & \heat{36.47} \\
$M_1$ & \seen{\textbf{45.92}} & 38.71 & 39.58 & 35.71 & 33.70 & 33.75 & 34.57 & 33.70 & \heat{37.15} \\
$M_2$ & \seen{45.92} & \seen{\textbf{44.46}} & 36.67 & 36.90 & 31.52 & 31.25 & 30.86 & 33.70 & \heat{36.67} \\
$M_3$ & \seen{51.02} & \seen{44.09} & \seen{\textbf{39.58}} & 40.67 & 33.70 & 30.00 & 33.33 & 27.17 & \heat{37.73} \\
$M_4$ & \seen{45.92} & \seen{45.16} & \seen{39.58} & \seen{\textbf{41.67}} & 31.70 & 33.75 & 33.33 & 31.52 & \heat{38.01} \\
$M_5$ & \seen{46.94} & \seen{38.71} & \seen{38.54} & \seen{41.67} & \seen{\textbf{34.43}} & 31.00 & 33.33 & 27.17 & \heat{37.66} \\
$M_6$ & \seen{46.94} & \seen{38.71} & \seen{39.46} & \seen{41.67} & \seen{34.43} & \seen{\textbf{35.00}} & 29.63 & 30.43 & \heat{37.23} \\
$M_7$ & \seen{47.96} & \seen{40.86} & \seen{37.38} & \seen{45.24} & \seen{33.70} & \seen{32.50} & \seen{\textbf{32.10}} & 28.26 & \heat{37.41} \\
$M_8$ & \seen{50.00} & \seen{41.94} & \seen{39.46} & \seen{46.43} & \seen{34.78} & \seen{34.50} & \seen{33.33} & \seen{\textbf{31.52}} & \heat{39.17} \\
\midrule
\multicolumn{10}{c}{\textbf{TriviaQA -- Qwen3-8B}} \\*
\midrule
\textbf{Eval ckpt} & $S_1$ (77) & $S_2$ (88) & $S_3$ (93) & $S_4$ (71) & $S_5$ (81) & $S_6$ (80) & $S_7$ (80) & $S_8$ (96) & Overall (666) \\*
\midrule
\setrange{41.12}{45.46}%
$M_0$ & 37.45 & 40.91 & 35.48 & 39.44 & 34.57 & 41.25 & 48.75 & 50.00 & \heat{41.12} \\
$M_1$ & \seen{\textbf{40.36}} & 39.91 & 41.94 & 38.03 & 33.33 & 40.00 & 43.75 & 50.00 & \heat{41.17} \\
$M_2$ & \seen{46.75} & \seen{\textbf{42.91}} & 38.94 & 40.66 & 40.74 & 41.25 & 48.75 & 46.88 & \heat{43.34} \\
$M_3$ & \seen{41.66} & \seen{43.18} & \seen{\textbf{41.48}} & 40.66 & 38.27 & 40.00 & 50.00 & 43.75 & \heat{42.42} \\
$M_4$ & \seen{38.96} & \seen{43.18} & \seen{41.94} & \seen{\textbf{47.89}} & 37.04 & 41.25 & 50.00 & 46.88 & \heat{43.40} \\
$M_5$ & \seen{38.96} & \seen{44.32} & \seen{40.86} & \seen{40.85} & \seen{\textbf{39.04}} & 40.00 & 46.25 & 46.88 & \heat{42.49} \\
$M_6$ & \seen{49.35} & \seen{42.91} & \seen{43.01} & \seen{46.48} & \seen{37.80} & \seen{\textbf{41.25}} & 45.00 & 55.21 & \heat{45.24} \\
$M_7$ & \seen{40.26} & \seen{45.45} & \seen{43.71} & \seen{40.85} & \seen{40.74} & \seen{43.75} & \seen{\textbf{49.75}} & 48.96 & \heat{44.36} \\
$M_8$ & \seen{42.86} & \seen{42.91} & \seen{43.78} & \seen{45.07} & \seen{39.80} & \seen{45.00} & \seen{51.50} & \seen{\textbf{51.92}} & \heat{45.46} \\
\midrule
\multicolumn{10}{c}{\textbf{TriviaQA -- Qwen3-14B}} \\*
\midrule
\textbf{Eval ckpt} & $S_1$ (78) & $S_2$ (67) & $S_3$ (81) & $S_4$ (78) & $S_5$ (61) & $S_6$ (77) & $S_7$ (84) & $S_8$ (66) & Overall (592) \\*
\midrule
\setrange{49.02}{53.50}%
$M_0$ & 41.03 & 55.22 & 44.44 & 51.85 & 53.38 & 53.25 & 52.76 & 46.97 & \heat{49.42} \\
$M_1$ & \seen{\textbf{47.44}} & 50.75 & 44.44 & 52.56 & 54.10 & 54.55 & 47.62 & 45.45 & \heat{49.49} \\
$M_2$ & \seen{46.15} & \seen{\textbf{50.75}} & 43.21 & 51.28 & 56.38 & 49.35 & 51.19 & 50.48 & \heat{49.61} \\
$M_3$ & \seen{44.87} & \seen{53.73} & \seen{\textbf{41.98}} & 50.56 & 56.38 & 51.95 & 47.62 & 53.03 & \heat{49.63} \\
$M_4$ & \seen{46.15} & \seen{53.73} & \seen{45.68} & \seen{\textbf{52.56}} & 56.38 & 49.35 & 51.19 & 50.00 & \heat{50.40} \\
$M_5$ & \seen{48.72} & \seen{51.25} & \seen{45.68} & \seen{53.85} & \seen{\textbf{59.02}} & 48.05 & 52.76 & 51.58 & \heat{51.13} \\
$M_6$ & \seen{45.87} & \seen{51.25} & \seen{45.68} & \seen{52.56} & \seen{57.38} & \seen{\textbf{50.65}} & 50.00 & 53.00 & \heat{50.52} \\
$M_7$ & \seen{50.00} & \seen{56.72} & \seen{41.98} & \seen{52.56} & \seen{59.02} & \seen{54.55} & \seen{\textbf{52.19}} & 50.00 & \heat{51.83} \\
$M_8$ & \seen{46.87} & \seen{55.22} & \seen{45.68} & \seen{53.85} & \seen{63.93} & \seen{58.44} & \seen{53.38} & \seen{\textbf{53.48}} & \heat{53.50} \\
\midrule
\multicolumn{10}{c}{\textbf{TriviaQA -- Qwen3-30B-A3B}} \\*
\midrule
\textbf{Eval ckpt} & $S_1$ (77) & $S_2$ (61) & $S_3$ (69) & $S_4$ (76) & $S_5$ (71) & $S_6$ (55) & $S_7$ (72) & $S_8$ (71) & Overall (552) \\*
\midrule
\setrange{44.95}{48.26}%
$M_0$ & 41.86 & 30.98 & 42.03 & 43.42 & 43.66 & 41.82 & 51.39 & 61.97 & \heat{44.95} \\
$M_1$ & \seen{\textbf{46.75}} & 32.62 & 39.13 & 47.37 & 40.85 & 36.36 & 54.17 & 63.38 & \heat{45.63} \\
$M_2$ & \seen{42.86} & \seen{\textbf{39.34}} & 38.23 & 46.05 & 42.25 & 38.18 & 52.78 & 61.97 & \heat{45.64} \\
$M_3$ & \seen{45.45} & \seen{36.07} & \seen{\textbf{42.13}} & 46.05 & 46.48 & 47.27 & 51.39 & 59.15 & \heat{46.93} \\
$M_4$ & \seen{46.75} & \seen{36.07} & \seen{36.33} & \seen{\textbf{51.32}} & 44.70 & 39.18 & 52.00 & 63.15 & \heat{46.67} \\
$M_5$ & \seen{48.05} & \seen{42.62} & \seen{45.03} & \seen{52.63} & \seen{\textbf{47.89}} & 38.18 & 49.17 & 59.01 & \heat{48.25} \\
$M_6$ & \seen{42.86} & \seen{39.34} & \seen{42.13} & \seen{47.05} & \seen{46.48} & \seen{\textbf{42.13}} & 51.78 & 61.97 & \heat{46.97} \\
$M_7$ & \seen{41.56} & \seen{36.07} & \seen{40.68} & \seen{47.05} & \seen{53.52} & \seen{42.13} & \seen{\textbf{54.17}} & 61.97 & \heat{47.46} \\
$M_8$ & \seen{48.05} & \seen{34.43} & \seen{42.13} & \seen{47.37} & \seen{45.07} & \seen{42.13} & \seen{56.94} & \seen{\textbf{66.38}} & \heat{48.26} \\
\midrule
\multicolumn{10}{c}{\textbf{TriviaQA -- Olmo-3-7B-Think}} \\*
\midrule
\textbf{Eval ckpt} & $S_1$ (127) & $S_2$ (131) & $S_3$ (133) & $S_4$ (116) & $S_5$ (118) & $S_6$ (107) & $S_7$ (124) & $S_8$ (123) & Overall (979) \\*
\midrule
\setrange{36.51}{43.20}%
$M_0$ & 42.88 & 42.75 & 33.08 & 40.52 & 34.75 & 35.51 & 29.84 & 32.52 & \heat{36.51} \\
$M_1$ & \seen{\textbf{48.16}} & 46.56 & 36.09 & 35.34 & 38.14 & 40.19 & 36.29 & 32.52 & \heat{39.24} \\
$M_2$ & \seen{48.03} & \seen{\textbf{49.62}} & 31.58 & 41.38 & 33.90 & 36.45 & 37.90 & 31.71 & \heat{38.92} \\
$M_3$ & \seen{48.03} & \seen{49.62} & \seen{\textbf{38.35}} & 44.41 & 32.20 & 40.19 & 39.52 & 31.71 & \heat{40.61} \\
$M_4$ & \seen{47.24} & \seen{49.62} & \seen{34.59} & \seen{\textbf{48.28}} & 33.59 & 39.25 & 41.13 & 28.46 & \heat{40.31} \\
$M_5$ & \seen{50.39} & \seen{50.38} & \seen{39.85} & \seen{48.28} & \seen{\textbf{35.59}} & 36.45 & 39.52 & 30.08 & \heat{41.47} \\
$M_6$ & \seen{54.33} & \seen{53.44} & \seen{36.09} & \seen{47.41} & \seen{33.90} & \seen{\textbf{39.25}} & 38.94 & 34.15 & \heat{42.32} \\
$M_7$ & \seen{46.46} & \seen{46.56} & \seen{38.35} & \seen{47.41} & \seen{35.59} & \seen{42.99} & \seen{\textbf{40.29}} & 30.15 & \heat{42.96} \\
$M_8$ & \seen{48.03} & \seen{49.62} & \seen{39.85} & \seen{50.86} & \seen{35.59} & \seen{42.99} & \seen{43.55} & \seen{\textbf{34.89}} & \heat{43.20} \\
\midrule
\multicolumn{10}{c}{\textbf{TriviaQA -- OLMoE-1B-7B}} \\*
\midrule
\textbf{Eval ckpt} & $S_1$ (135) & $S_2$ (123) & $S_3$ (136) & $S_4$ (121) & $S_5$ (125) & $S_6$ (111) & $S_7$ (128) & $S_8$ (124) & Overall (1003) \\*
\midrule
\setrange{34.88}{39.78}%
$M_0$ & 37.22 & 33.77 & 27.88 & 36.15 & 34.20 & 40.63 & 39.06 & 31.06 & \heat{34.88} \\
$M_1$ & \seen{\textbf{42.22}} & 35.77 & 30.15 & 41.32 & 37.60 & 40.63 & 41.41 & 34.68 & \heat{37.90} \\
$M_2$ & \seen{37.04} & \seen{\textbf{38.21}} & 27.21 & 37.19 & 41.60 & 37.84 & 40.62 & 38.71 & \heat{37.19} \\
$M_3$ & \seen{37.04} & \seen{35.77} & \seen{\textbf{33.09}} & 36.67 & 40.80 & 35.14 & 42.19 & 38.71 & \heat{37.43} \\
$M_4$ & \seen{36.81} & \seen{39.02} & \seen{37.50} & \seen{\textbf{36.36}} & 32.00 & 32.43 & 41.41 & 36.06 & \heat{37.06} \\
$M_5$ & \seen{36.07} & \seen{40.65} & \seen{36.76} & \seen{34.71} & \seen{\textbf{36.00}} & 34.04 & 47.66 & 33.87 & \heat{37.53} \\
$M_6$ & \seen{41.48} & \seen{39.84} & \seen{40.44} & \seen{41.32} & \seen{39.20} & \seen{\textbf{36.04}} & 41.41 & 35.71 & \heat{39.51} \\
$M_7$ & \seen{34.81} & \seen{41.46} & \seen{37.50} & \seen{39.67} & \seen{40.80} & \seen{35.14} & \seen{\textbf{49.00}} & 32.26 & \heat{38.86} \\
$M_8$ & \seen{36.30} & \seen{41.46} & \seen{36.76} & \seen{37.19} & \seen{40.00} & \seen{36.04} & \seen{49.22} & \seen{\textbf{45.97}} & \heat{40.38} \\
\end{longtable}
\endgroup
 
\begingroup
\renewcommand{\arraystretch}{0.45}
\setlength{\LTcapwidth}{\textwidth}
\captionsetup{font=normalsize}
\footnotesize
\setlength{\tabcolsep}{3pt}
\begin{longtable}{cccccccccc}
\caption{Scores on \texttt{HalluSet-heldout}, binned by hallucination point. Rows denote checkpoints. $S_t$ denotes questions whose corresponding hallucinations were generated between checkpoints $M_{t-1}$ and $M_t$ (counts in parentheses). Bold cells denote the first evaluated checkpoint after the hallucination point for each column's facts. Scores generally increase initially due to RL-induced distribution sharpening, but then become steady because there is no reward signal.}

\label{tab:per_bucket_scores_neutral}\\
\toprule
\endfirsthead
\multicolumn{10}{c}{\tablename~\thetable{} -- \textit{continued from previous page}}\\
\toprule
\endhead
\midrule
\multicolumn{10}{r}{\textit{continued on next page}}\\
\endfoot
\bottomrule
\endlastfoot
\multicolumn{10}{c}{\textbf{QuALITY -- Qwen3-4B}} \\*
\midrule
\textbf{Eval ckpt} & $S_1$ (125) & $S_2$ (125) & $S_3$ (125) & $S_4$ (125) & $S_5$ (125) & $S_6$ (125) & $S_7$ (125) & $S_8$ (125) & Overall (1000) \\*
\midrule
\setrange{69.40}{79.10}%
$M_0$ & 66.40 & 66.40 & 60.00 & 69.60 & 76.80 & 70.40 & 76.00 & 69.60 & \heat{69.40} \\
$M_1$ & \seen{\textbf{68.00}} & 75.20 & 66.40 & 72.80 & 82.40 & 74.40 & 78.40 & 75.20 & \heat{74.10} \\
$M_2$ & \seen{70.40} & \seen{\textbf{76.80}} & 72.80 & 80.80 & 80.80 & 77.60 & 83.20 & 75.20 & \heat{77.20} \\
$M_3$ & \seen{72.00} & \seen{79.20} & \seen{\textbf{71.20}} & 80.00 & 84.80 & 76.80 & 81.60 & 78.40 & \heat{78.00} \\
$M_4$ & \seen{72.80} & \seen{78.40} & \seen{73.60} & \seen{\textbf{79.20}} & 82.40 & 79.20 & 86.40 & 80.80 & \heat{79.10} \\
$M_5$ & \seen{72.00} & \seen{78.40} & \seen{72.00} & \seen{76.80} & \seen{\textbf{83.20}} & 76.80 & 84.00 & 78.40 & \heat{77.70} \\
$M_6$ & \seen{72.80} & \seen{76.80} & \seen{68.00} & \seen{77.60} & \seen{84.00} & \seen{\textbf{78.40}} & 86.40 & 80.80 & \heat{78.10} \\
$M_7$ & \seen{73.60} & \seen{76.80} & \seen{68.80} & \seen{79.20} & \seen{84.00} & \seen{79.20} & \seen{\textbf{87.20}} & 83.20 & \heat{79.00} \\
$M_8$ & \seen{72.80} & \seen{78.40} & \seen{68.80} & \seen{79.20} & \seen{84.00} & \seen{80.80} & \seen{84.80} & \seen{\textbf{83.20}} & \heat{79.00} \\
\midrule
\multicolumn{10}{c}{\textbf{QuALITY -- Qwen3-8B}} \\*
\midrule
\textbf{Eval ckpt} & $S_1$ (125) & $S_2$ (125) & $S_3$ (125) & $S_4$ (125) & $S_5$ (125) & $S_6$ (125) & $S_7$ (125) & $S_8$ (125) & Overall (1000) \\*
\midrule
\setrange{73.25}{84.70}%
$M_0$ & 68.80 & 78.00 & 70.40 & 72.80 & 69.60 & 74.40 & 78.40 & 73.60 & \heat{73.25} \\
$M_1$ & \seen{\textbf{64.80}} & 85.60 & 76.80 & 77.60 & 77.60 & 77.60 & 79.20 & 76.80 & \heat{77.00} \\
$M_2$ & \seen{69.60} & \seen{\textbf{82.40}} & 83.20 & 78.40 & 81.60 & 76.80 & 83.20 & 80.00 & \heat{79.40} \\
$M_3$ & \seen{65.60} & \seen{85.60} & \seen{\textbf{84.00}} & 84.80 & 82.40 & 76.80 & 81.60 & 80.00 & \heat{80.10} \\
$M_4$ & \seen{74.40} & \seen{83.20} & \seen{85.60} & \seen{\textbf{86.40}} & 83.20 & 78.40 & 84.00 & 84.00 & \heat{82.40} \\
$M_5$ & \seen{74.40} & \seen{83.20} & \seen{85.60} & \seen{84.00} & \seen{\textbf{86.40}} & 81.60 & 85.60 & 84.80 & \heat{83.20} \\
$M_6$ & \seen{76.00} & \seen{85.60} & \seen{86.40} & \seen{84.80} & \seen{85.60} & \seen{\textbf{81.60}} & 88.00 & 86.40 & \heat{84.30} \\
$M_7$ & \seen{74.40} & \seen{86.40} & \seen{88.00} & \seen{86.40} & \seen{86.40} & \seen{81.60} & \seen{\textbf{86.40}} & 88.00 & \heat{84.70} \\
$M_8$ & \seen{74.40} & \seen{84.80} & \seen{84.00} & \seen{85.60} & \seen{85.60} & \seen{80.80} & \seen{84.80} & \seen{\textbf{85.60}} & \heat{83.20} \\
\midrule
\multicolumn{10}{c}{\textbf{QuALITY -- Qwen3-14B}} \\*
\midrule
\textbf{Eval ckpt} & $S_1$ (125) & $S_2$ (125) & $S_3$ (125) & $S_4$ (125) & $S_5$ (125) & $S_6$ (125) & $S_7$ (125) & $S_8$ (125) & Overall (1000) \\*
\midrule
\setrange{82.00}{85.70}%
$M_0$ & 76.00 & 80.80 & 85.60 & 80.00 & 84.00 & 80.80 & 84.80 & 84.00 & \heat{82.00} \\
$M_1$ & \seen{\textbf{78.40}} & 82.40 & 85.60 & 84.00 & 86.40 & 81.60 & 83.20 & 86.40 & \heat{83.50} \\
$M_2$ & \seen{80.00} & \seen{\textbf{81.60}} & 84.00 & 82.40 & 86.40 & 85.60 & 85.60 & 87.20 & \heat{84.10} \\
$M_3$ & \seen{78.40} & \seen{84.80} & \seen{\textbf{85.60}} & 84.00 & 86.40 & 82.40 & 85.60 & 87.20 & \heat{84.30} \\
$M_4$ & \seen{80.00} & \seen{83.20} & \seen{86.40} & \seen{\textbf{83.20}} & 90.40 & 84.00 & 84.00 & 88.00 & \heat{84.90} \\
$M_5$ & \seen{80.80} & \seen{83.20} & \seen{85.60} & \seen{86.40} & \seen{\textbf{89.60}} & 82.40 & 86.40 & 88.80 & \heat{85.40} \\
$M_6$ & \seen{83.20} & \seen{83.20} & \seen{84.80} & \seen{86.40} & \seen{88.80} & \seen{\textbf{82.40}} & 87.20 & 89.60 & \heat{85.70} \\
$M_7$ & \seen{80.80} & \seen{84.00} & \seen{84.80} & \seen{84.80} & \seen{84.00} & \seen{84.80} & \seen{\textbf{86.40}} & 88.00 & \heat{84.70} \\
$M_8$ & \seen{80.80} & \seen{85.60} & \seen{83.20} & \seen{81.60} & \seen{84.00} & \seen{84.00} & \seen{86.40} & \seen{\textbf{88.80}} & \heat{84.30} \\
\midrule
\multicolumn{10}{c}{\textbf{QuALITY -- Qwen3-30B-A3B}} \\*
\midrule
\textbf{Eval ckpt} & $S_1$ (125) & $S_2$ (125) & $S_3$ (125) & $S_4$ (125) & $S_5$ (125) & $S_6$ (125) & $S_7$ (125) & $S_8$ (125) & Overall (1000) \\*
\midrule
\setrange{75.70}{83.30}%
$M_0$ & 75.20 & 76.00 & 76.00 & 80.00 & 74.40 & 71.20 & 73.60 & 79.20 & \heat{75.70} \\
$M_1$ & \seen{\textbf{72.00}} & 80.00 & 79.20 & 80.80 & 76.00 & 76.00 & 81.60 & 83.20 & \heat{78.60} \\
$M_2$ & \seen{75.20} & \seen{\textbf{80.80}} & 84.80 & 84.00 & 79.20 & 76.80 & 78.40 & 84.00 & \heat{80.40} \\
$M_3$ & \seen{79.20} & \seen{81.60} & \seen{\textbf{86.40}} & 84.00 & 80.80 & 80.00 & 80.00 & 86.40 & \heat{82.30} \\
$M_4$ & \seen{80.00} & \seen{80.80} & \seen{84.80} & \seen{\textbf{84.00}} & 80.80 & 79.20 & 80.00 & 86.40 & \heat{82.00} \\
$M_5$ & \seen{79.20} & \seen{83.20} & \seen{86.40} & \seen{84.00} & \seen{\textbf{79.20}} & 80.00 & 83.20 & 84.80 & \heat{82.50} \\
$M_6$ & \seen{81.60} & \seen{83.20} & \seen{85.60} & \seen{84.80} & \seen{78.40} & \seen{\textbf{82.40}} & 84.80 & 85.60 & \heat{83.30} \\
$M_7$ & \seen{80.80} & \seen{84.80} & \seen{85.60} & \seen{83.20} & \seen{83.20} & \seen{79.20} & \seen{\textbf{83.20}} & 86.40 & \heat{83.30} \\
$M_8$ & \seen{80.80} & \seen{84.80} & \seen{84.80} & \seen{84.00} & \seen{81.60} & \seen{80.00} & \seen{84.00} & \seen{\textbf{84.80}} & \heat{83.10} \\
\midrule
\multicolumn{10}{c}{\textbf{QuALITY -- Olmo-3-7B-Think}} \\*
\midrule
\textbf{Eval ckpt} & $S_1$ (125) & $S_2$ (125) & $S_3$ (125) & $S_4$ (125) & $S_5$ (125) & $S_6$ (125) & $S_7$ (125) & $S_8$ (125) & Overall (1000) \\*
\midrule
\setrange{77.80}{85.40}%
$M_0$ & 80.80 & 80.80 & 76.00 & 78.40 & 79.20 & 72.00 & 75.20 & 80.00 & \heat{77.80} \\
$M_1$ & \seen{\textbf{80.00}} & 84.00 & 76.00 & 83.20 & 80.00 & 78.40 & 85.60 & 85.60 & \heat{81.60} \\
$M_2$ & \seen{79.20} & \seen{\textbf{82.40}} & 84.80 & 85.60 & 84.00 & 81.60 & 84.00 & 84.80 & \heat{83.30} \\
$M_3$ & \seen{83.20} & \seen{84.80} & \seen{\textbf{80.00}} & 84.00 & 81.60 & 79.20 & 86.40 & 84.80 & \heat{83.00} \\
$M_4$ & \seen{84.00} & \seen{84.80} & \seen{85.60} & \seen{\textbf{84.80}} & 84.00 & 79.20 & 86.40 & 83.20 & \heat{84.00} \\
$M_5$ & \seen{83.20} & \seen{87.20} & \seen{80.80} & \seen{85.60} & \seen{\textbf{84.00}} & 84.00 & 88.00 & 82.40 & \heat{84.40} \\
$M_6$ & \seen{84.00} & \seen{85.60} & \seen{84.00} & \seen{85.60} & \seen{85.60} & \seen{\textbf{81.60}} & 89.60 & 84.80 & \heat{85.10} \\
$M_7$ & \seen{84.80} & \seen{84.80} & \seen{85.60} & \seen{84.80} & \seen{84.00} & \seen{84.00} & \seen{\textbf{88.80}} & 86.40 & \heat{85.40} \\
$M_8$ & \seen{85.60} & \seen{85.60} & \seen{87.20} & \seen{84.80} & \seen{85.60} & \seen{81.60} & \seen{88.00} & \seen{\textbf{84.80}} & \heat{85.40} \\
\midrule
\multicolumn{10}{c}{\textbf{QuALITY -- OLMoE-1B-7B}} \\*
\midrule
\textbf{Eval ckpt} & $S_1$ (125) & $S_2$ (125) & $S_3$ (125) & $S_4$ (125) & $S_5$ (125) & $S_6$ (125) & $S_7$ (125) & $S_8$ (125) & Overall (1000) \\*
\midrule
\setrange{58.80}{67.80}%
$M_0$ & 55.20 & 67.20 & 52.80 & 53.60 & 64.00 & 56.00 & 61.60 & 60.00 & \heat{58.80} \\
$M_1$ & \seen{\textbf{52.00}} & 69.60 & 54.40 & 56.80 & 68.80 & 61.00 & 61.60 & 64.00 & \heat{61.03} \\
$M_2$ & \seen{51.20} & \seen{\textbf{69.60}} & 62.40 & 61.60 & 70.40 & 64.00 & 68.00 & 64.00 & \heat{63.90} \\
$M_3$ & \seen{55.20} & \seen{68.80} & \seen{\textbf{62.40}} & 67.20 & 72.80 & 62.40 & 69.60 & 69.60 & \heat{66.00} \\
$M_4$ & \seen{54.40} & \seen{70.40} & \seen{59.20} & \seen{\textbf{64.80}} & 72.00 & 64.80 & 69.60 & 69.60 & \heat{65.60} \\
$M_5$ & \seen{56.80} & \seen{71.20} & \seen{64.80} & \seen{69.60} & \seen{\textbf{71.20}} & 67.20 & 71.20 & 68.00 & \heat{67.50} \\
$M_6$ & \seen{59.20} & \seen{69.60} & \seen{65.60} & \seen{66.40} & \seen{72.00} & \seen{\textbf{68.80}} & 71.20 & 69.60 & \heat{67.80} \\
$M_7$ & \seen{56.80} & \seen{72.80} & \seen{64.80} & \seen{68.80} & \seen{72.80} & \seen{65.60} & \seen{\textbf{72.00}} & 68.00 & \heat{67.70} \\
$M_8$ & \seen{56.00} & \seen{70.40} & \seen{60.00} & \seen{68.00} & \seen{72.80} & \seen{68.00} & \seen{73.60} & \seen{\textbf{68.80}} & \heat{67.20} \\
\midrule
\multicolumn{10}{c}{\textbf{TriviaQA -- Qwen3-4B}} \\*
\midrule
\textbf{Eval ckpt} & $S_1$ (115) & $S_2$ (99) & $S_3$ (106) & $S_4$ (111) & $S_5$ (102) & $S_6$ (110) & $S_7$ (108) & $S_8$ (110) & Overall (861) \\*
\midrule
\setrange{36.93}{39.95}%
$M_0$ & 37.39 & 35.35 & 43.40 & 38.74 & 38.24 & 31.82 & 32.41 & 38.18 & \heat{36.93} \\
$M_1$ & \seen{\textbf{39.13}} & 35.35 & 46.23 & 42.34 & 41.18 & 37.27 & 37.96 & 36.36 & \heat{39.49} \\
$M_2$ & \seen{40.87} & \seen{\textbf{34.34}} & 48.11 & 41.44 & 41.18 & 39.09 & 34.26 & 36.36 & \heat{39.49} \\
$M_3$ & \seen{39.13} & \seen{35.35} & \seen{\textbf{45.28}} & 39.64 & 45.10 & 37.27 & 39.81 & 37.27 & \heat{39.84} \\
$M_4$ & \seen{40.87} & \seen{33.33} & \seen{45.28} & \seen{\textbf{39.64}} & 45.10 & 38.18 & 39.81 & 36.36 & \heat{39.84} \\
$M_5$ & \seen{38.26} & \seen{36.36} & \seen{48.11} & \seen{42.34} & \seen{\textbf{45.10}} & 38.18 & 36.11 & 35.45 & \heat{39.95} \\
$M_6$ & \seen{39.13} & \seen{35.35} & \seen{46.23} & \seen{40.54} & \seen{40.20} & \seen{\textbf{38.18}} & 37.96 & 33.64 & \heat{38.91} \\
$M_7$ & \seen{39.13} & \seen{38.38} & \seen{48.11} & \seen{39.64} & \seen{41.18} & \seen{40.91} & \seen{\textbf{35.19}} & 37.27 & \heat{39.95} \\
$M_8$ & \seen{38.26} & \seen{35.35} & \seen{40.57} & \seen{40.54} & \seen{39.22} & \seen{40.91} & \seen{35.19} & \seen{\textbf{35.45}} & \heat{38.21} \\
\midrule
\multicolumn{10}{c}{\textbf{TriviaQA -- Qwen3-8B}} \\*
\midrule
\textbf{Eval ckpt} & $S_1$ (109) & $S_2$ (100) & $S_3$ (97) & $S_4$ (98) & $S_5$ (90) & $S_6$ (106) & $S_7$ (101) & $S_8$ (94) & Overall (795) \\*
\midrule
\setrange{43.27}{46.41}%
$M_0$ & 51.38 & 41.00 & 45.36 & 40.82 & 37.78 & 40.57 & 39.60 & 48.93 & \heat{43.27} \\
$M_1$ & \seen{\textbf{52.29}} & 44.00 & 47.42 & 41.84 & 38.89 & 44.34 & 40.59 & 50.00 & \heat{45.03} \\
$M_2$ & \seen{50.46} & \seen{\textbf{41.00}} & 47.42 & 43.88 & 41.11 & 44.34 & 45.54 & 53.19 & \heat{45.91} \\
$M_3$ & \seen{47.71} & \seen{41.00} & \seen{\textbf{48.45}} & 41.84 & 40.00 & 41.51 & 41.58 & 53.19 & \heat{44.40} \\
$M_4$ & \seen{52.29} & \seen{42.00} & \seen{48.45} & \seen{\textbf{41.84}} & 40.00 & 45.28 & 42.57 & 56.38 & \heat{46.16} \\
$M_5$ & \seen{51.38} & \seen{42.00} & \seen{48.45} & \seen{40.82} & \seen{\textbf{40.00}} & 47.17 & 46.53 & 53.19 & \heat{46.29} \\
$M_6$ & \seen{48.62} & \seen{41.00} & \seen{48.45} & \seen{42.86} & \seen{37.78} & \seen{\textbf{45.28}} & 43.56 & 54.26 & \heat{45.28} \\
$M_7$ & \seen{51.38} & \seen{41.00} & \seen{44.33} & \seen{41.84} & \seen{43.33} & \seen{45.28} & \seen{\textbf{47.52}} & 55.32 & \heat{46.29} \\
$M_8$ & \seen{48.62} & \seen{40.00} & \seen{45.36} & \seen{41.84} & \seen{37.78} & \seen{42.45} & \seen{41.58} & \seen{\textbf{55.32}} & \heat{44.15} \\
\midrule
\multicolumn{10}{c}{\textbf{TriviaQA -- Qwen3-14B}} \\*
\midrule
\textbf{Eval ckpt} & $S_1$ (95) & $S_2$ (95) & $S_3$ (88) & $S_4$ (91) & $S_5$ (89) & $S_6$ (100) & $S_7$ (99) & $S_8$ (95) & Overall (752) \\*
\midrule
\setrange{44.95}{47.74}%
$M_0$ & 45.26 & 34.74 & 51.14 & 47.25 & 35.96 & 50.00 & 45.45 & 49.47 & \heat{44.95} \\
$M_1$ & \seen{\textbf{46.32}} & 38.95 & 46.59 & 50.55 & 39.33 & 51.00 & 51.52 & 54.74 & \heat{47.48} \\
$M_2$ & \seen{46.32} & \seen{\textbf{38.95}} & 47.73 & 47.25 & 35.96 & 52.00 & 52.53 & 50.53 & \heat{46.55} \\
$M_3$ & \seen{48.42} & \seen{40.00} & \seen{\textbf{47.73}} & 50.55 & 35.96 & 51.00 & 49.49 & 50.53 & \heat{46.81} \\
$M_4$ & \seen{46.32} & \seen{34.74} & \seen{55.68} & \seen{\textbf{47.25}} & 39.33 & 50.00 & 52.53 & 50.53 & \heat{47.07} \\
$M_5$ & \seen{46.32} & \seen{34.74} & \seen{51.14} & \seen{49.45} & \seen{\textbf{39.33}} & 51.00 & 50.51 & 52.63 & \heat{46.94} \\
$M_6$ & \seen{54.74} & \seen{37.89} & \seen{46.59} & \seen{46.15} & \seen{37.08} & \seen{\textbf{50.00}} & 50.51 & 50.53 & \heat{46.81} \\
$M_7$ & \seen{47.37} & \seen{36.84} & \seen{56.82} & \seen{47.25} & \seen{37.08} & \seen{51.00} & \seen{\textbf{54.55}} & 50.53 & \heat{47.74} \\
$M_8$ & \seen{48.42} & \seen{36.84} & \seen{52.27} & \seen{43.96} & \seen{39.33} & \seen{51.00} & \seen{49.49} & \seen{\textbf{50.53}} & \heat{46.54} \\
\midrule
\multicolumn{10}{c}{\textbf{TriviaQA -- Qwen3-30B-A3B}} \\*
\midrule
\textbf{Eval ckpt} & $S_1$ (97) & $S_2$ (93) & $S_3$ (90) & $S_4$ (91) & $S_5$ (87) & $S_6$ (94) & $S_7$ (89) & $S_8$ (89) & Overall (730) \\*
\midrule
\setrange{44.66}{46.44}%
$M_0$ & 44.33 & 49.46 & 34.44 & 40.66 & 50.57 & 52.13 & 42.70 & 47.19 & \heat{45.20} \\
$M_1$ & \seen{\textbf{42.27}} & 50.54 & 36.67 & 42.86 & 47.13 & 53.19 & 43.82 & 47.19 & \heat{45.48} \\
$M_2$ & \seen{47.42} & \seen{\textbf{46.24}} & 34.44 & 41.76 & 50.57 & 53.19 & 42.70 & 52.81 & \heat{46.16} \\
$M_3$ & \seen{46.39} & \seen{48.39} & \seen{\textbf{36.67}} & 41.76 & 49.43 & 52.13 & 46.07 & 50.56 & \heat{46.44} \\
$M_4$ & \seen{47.42} & \seen{50.54} & \seen{34.44} & \seen{\textbf{42.86}} & 48.28 & 54.26 & 43.82 & 47.19 & \heat{46.16} \\
$M_5$ & \seen{43.30} & \seen{48.39} & \seen{37.78} & \seen{38.46} & \seen{\textbf{50.57}} & 52.13 & 46.07 & 48.31 & \heat{45.62} \\
$M_6$ & \seen{43.30} & \seen{48.39} & \seen{38.89} & \seen{41.76} & \seen{49.43} & \seen{\textbf{53.19}} & 43.82 & 48.31 & \heat{45.89} \\
$M_7$ & \seen{44.33} & \seen{50.54} & \seen{37.78} & \seen{40.66} & \seen{49.43} & \seen{52.13} & \seen{\textbf{42.70}} & 48.31 & \heat{45.75} \\
$M_8$ & \seen{46.39} & \seen{49.46} & \seen{34.44} & \seen{37.36} & \seen{45.98} & \seen{54.26} & \seen{42.70} & \seen{\textbf{46.07}} & \heat{44.66} \\
\midrule
\multicolumn{10}{c}{\textbf{TriviaQA -- Olmo-3-7B-Think}} \\*
\midrule
\textbf{Eval ckpt} & $S_1$ (104) & $S_2$ (115) & $S_3$ (111) & $S_4$ (112) & $S_5$ (105) & $S_6$ (107) & $S_7$ (103) & $S_8$ (105) & Overall (862) \\*
\midrule
\setrange{34.69}{43.85}%
$M_0$ & 35.58 & 33.04 & 30.63 & 36.61 & 34.29 & 40.18 & 30.10 & 37.14 & \heat{34.69} \\
$M_1$ & \seen{\textbf{40.38}} & 36.52 & 36.94 & 42.86 & 40.00 & 45.79 & 36.89 & 42.86 & \heat{40.25} \\
$M_2$ & \seen{37.50} & \seen{\textbf{37.39}} & 35.14 & 46.43 & 40.00 & 42.06 & 41.75 & 42.86 & \heat{40.37} \\
$M_3$ & \seen{42.31} & \seen{40.87} & \seen{\textbf{35.14}} & 44.64 & 42.86 & 42.06 & 40.78 & 43.81 & \heat{41.53} \\
$M_4$ & \seen{41.35} & \seen{38.26} & \seen{37.84} & \seen{\textbf{49.11}} & 44.76 & 44.86 & 36.89 & 48.57 & \heat{42.69} \\
$M_5$ & \seen{41.35} & \seen{39.13} & \seen{35.14} & \seen{47.32} & \seen{\textbf{45.71}} & 47.66 & 44.66 & 50.48 & \heat{43.85} \\
$M_6$ & \seen{37.50} & \seen{40.00} & \seen{36.94} & \seen{47.32} & \seen{45.71} & \seen{\textbf{44.86}} & 38.83 & 48.57 & \heat{42.46} \\
$M_7$ & \seen{38.46} & \seen{38.26} & \seen{33.33} & \seen{45.54} & \seen{43.81} & \seen{44.86} & \seen{\textbf{36.89}} & 48.57 & \heat{41.18} \\
$M_8$ & \seen{37.50} & \seen{40.00} & \seen{34.23} & \seen{47.32} & \seen{44.76} & \seen{44.86} & \seen{38.83} & \seen{\textbf{47.62}} & \heat{41.88} \\
\midrule
\multicolumn{10}{c}{\textbf{TriviaQA -- OLMoE-1B-7B}} \\*
\midrule
\textbf{Eval ckpt} & $S_1$ (98) & $S_2$ (95) & $S_3$ (92) & $S_4$ (92) & $S_5$ (102) & $S_6$ (105) & $S_7$ (99) & $S_8$ (90) & Overall (773) \\*
\midrule
\setrange{30.27}{33.38}%
$M_0$ & 31.63 & 32.63 & 26.09 & 21.74 & 30.39 & 34.29 & 29.29 & 35.55 & \heat{30.27} \\
$M_1$ & \seen{\textbf{34.69}} & 33.68 & 26.09 & 28.26 & 30.39 & 34.29 & 38.38 & 41.11 & \heat{33.38} \\
$M_2$ & \seen{31.63} & \seen{\textbf{34.74}} & 27.17 & 28.26 & 30.39 & 31.43 & 34.34 & 37.78 & \heat{31.95} \\
$M_3$ & \seen{33.67} & \seen{32.63} & \seen{\textbf{26.09}} & 26.09 & 34.31 & 34.29 & 35.35 & 37.78 & \heat{32.60} \\
$M_4$ & \seen{32.65} & \seen{33.68} & \seen{25.00} & \seen{\textbf{21.74}} & 33.33 & 36.19 & 39.39 & 38.89 & \heat{32.73} \\
$M_5$ & \seen{34.69} & \seen{30.53} & \seen{25.00} & \seen{25.00} & \seen{\textbf{38.24}} & 35.23 & 39.39 & 36.67 & \heat{33.25} \\
$M_6$ & \seen{34.69} & \seen{30.53} & \seen{26.09} & \seen{22.83} & \seen{34.31} & \seen{\textbf{33.33}} & 38.38 & 40.00 & \heat{32.60} \\
$M_7$ & \seen{33.67} & \seen{29.47} & \seen{20.65} & \seen{25.00} & \seen{38.24} & \seen{33.33} & \seen{\textbf{41.41}} & 36.67 & \heat{32.47} \\
$M_8$ & \seen{34.69} & \seen{29.47} & \seen{20.65} & \seen{22.83} & \seen{36.27} & \seen{34.29} & \seen{40.40} & \seen{\textbf{38.89}} & \heat{32.34} \\
\end{longtable}
\endgroup

\begin{table}[ht]
\caption{Scores on \texttt{HalluSet-train} for checkpoints saved at RL epochs 1.5 and 2. $S_t$ denotes the set of hallucinations generated between checkpoints $M_{t-1}$ and $M_t$. For each $S_t$, $M_{p(h)+1}$ denotes the first checkpoint after the hallucination point. Compared to $M_{p(h)+1}$, later checkpoints generally maintain or improve their scores as training progresses, i.e., \textbf{models do not forget hallucinations memorized early during training.}}

\label{tab:epoch2_scores}
\centering
\footnotesize
\setlength{\tabcolsep}{4pt}
\renewcommand{\arraystretch}{1.0}
\begin{tabular*}{\textwidth}{@{\extracolsep{\fill}}cccccccccc}
\toprule
\textbf{Model} & \textbf{Eval ckpt} & $S_1$ & $S_2$ & $S_3$ & $S_4$ & $S_5$ & $S_6$ & $S_7$ & $S_8$ \\
\midrule
\multicolumn{10}{c}{\textbf{QuALITY}} \\
\midrule
\multirow{3}{*}{Qwen3-4B}
  & $M_{p(h)+1}$& 69.85 & 70.38 & 74.16 & 73.80 & 76.13 & 78.20 & 71.35 & 80.00 \\
  & Epoch 1.5        & 76.72 & 79.94 & 76.78 & 75.55 & 74.13 & 79.62 & 77.08 & 79.59 \\
  & Epoch 2          & 75.43 & 76.48 & 75.28 & 76.93 & 77.77 & 80.67 & 76.88 & 81.05 \\
\midrule
\multirow{3}{*}{Qwen3-8B}
  & $M_{p(h)+1}$& 72.19 & 74.74 & 81.74 & 78.92 & 87.43 & 82.32 & 85.57 & 85.11 \\
  & Epoch 1.5        & 81.46 & 81.75 & 82.33 & 82.51 & 86.39 & 81.79 & 87.73 & 85.08 \\
  & Epoch 2          & 84.44 & 85.61 & 85.48 & 82.51 & 86.91 & 82.80 & 88.27 & 84.46 \\
\midrule
\multirow{3}{*}{Qwen3-14B}
  & $M_{p(h)+1}$& 85.92 & 82.20 & 81.87 & 86.67 & 83.67 & 86.33 & 90.72 & 80.57 \\
  & Epoch 1.5        & 87.00 & 82.20 & 80.87 & 88.89 & 82.31 & 89.21 & 87.22 & 82.14 \\
  & Epoch 2          & 87.53 & 82.63 & 85.79 & 88.33 & 85.03 & 87.05 & 86.57 & 81.25 \\
\midrule
\multirow{3}{*}{Qwen3-30B-A3B}
  & $M_{p(h)+1}$& 72.96 & 72.56 & 78.16 & 77.54 & 83.53 & 83.12 & 82.31 & 83.45 \\
  & Epoch 1.5        & 76.10 &	78.20 &	82.04 &	81.82 &	85.88 &	84.34 &	85.71 &	84.17 \\
  & Epoch 2          & 83.09 &	74.81 &	82.04 &	79.14 &	80.00 &	83.92 &	88.44 &	83.45 \\
\midrule
\multirow{3}{*}{Olmo-3-7B-Think}
  & $M_{p(h)+1}$& 78.70 & 81.33 & 83.93 & 86.59 & 83.20 & 85.27 & 85.97 & 85.59 \\
  & Epoch 1.5        &87.57	&84.49	&88.20&	86.96&	83.33&	83.72&	84.39&	85.59 \\
  & Epoch 2         &85.21	&86.09	&87.21&	88.04&	85.16&	84.78&	85.65&	85.59\\
\midrule
\multirow{3}{*}{OLMoE-1B-7B}
  & $M_{p(h)+1}$& 54.20 & 53.85 & 64.09 & 64.38 & 70.16 & 71.73 & 73.62 & 73.13 \\
  & Epoch 1.5        &64.78&	64.70&	62.54&	66.54&	71.89&	73.96&	68.27&	70.45 \\
  & Epoch 2          &67.75&	66.67&	76.47&	71.57&	77.38&	74.20&	70.47&	73.13 \\
\midrule
\multicolumn{10}{c}{\textbf{TriviaQA}} \\
\midrule
\multirow{3}{*}{Qwen3-4B}
  & $M_{p(h)+1}$& 45.92 & 44.46 & 39.58 & 41.67 & 34.43 & 35.00 & 32.10 & 31.52 \\
  & Epoch 1.5        & 46.94 & 42.78 & 40.62 & 46.43 & 34.70 & 35.00 & 34.57 & 31.52 \\
  & Epoch 2          & 47.04 & 41.94 & 42.71 & 44.90 & 35.87 & 36.25 & 35.93 & 32.43 \\
\midrule
\multirow{3}{*}{Qwen3-8B}
  & $M_{p(h)+1}$& 40.36 & 42.91 & 41.48 & 47.89 & 39.04 & 41.25 & 49.75 & 51.92 \\
  & Epoch 1.5        & 42.86 & 43.18 & 42.78 & 45.03 & 37.04 & 44.50 & 49.25 & 53.96 \\
  & Epoch 2          & 42.86 & 44.05 & 42.63 & 45.03 & 39.51 & 45.05 & 51.00 & 52.08 \\
\midrule
\multirow{3}{*}{Qwen3-14B}
  & $M_{p(h)+1}$& 47.44 & 50.75 & 41.98 & 52.56 & 59.02 & 50.65 & 52.19 & 53.48 \\
  & Epoch 1.5        & 47.87 & 52.24 & 43.21 & 51.00 & 58.38 & 50.65 & 52.43 & 52.52 \\
  & Epoch 2          & 48.72 & 55.22 & 48.15 & 52.56 & 58.38 & 53.25 & 51.81 & 53.35 \\
\midrule
\multirow{3}{*}{Qwen3-30B-A3B}
  & $M_{p(h)+1}$& 46.75 & 39.34 & 42.13 & 51.32 & 47.89 & 42.13 & 54.17 & 66.38 \\
  & Epoch 1.5        & 44.16	&38.43	&44.78	&49.05	&49.30	&43.18	&53.78	&66.38 \\
  & Epoch 2          & 48.05	&38.07	&42.03	&51.32	&49.30	&45.45	&51.61	&65.67 \\

\midrule
\multirow{3}{*}{Olmo-3-7B-Think}
  & $M_{p(h)+1}$& 48.16 & 49.62 & 38.35 & 48.28 & 35.59 & 39.25 & 40.29 & 34.89 \\
  & Epoch 1.5        & 47.24	 & 49.62	 & 36.84	 & 49.14	 & 35.59	 & 40.45	 & 42.29	 & 33.71\\
  & Epoch 2          & 48.09	 & 50.38	 & 38.35	 & 48.69	 & 35.59	 & 39.58	 & 42.29	 & 35.08 \\

\midrule
\multirow{3}{*}{OLMoE-1B-7B}
  & $M_{p(h)+1}$& 42.22 & 38.21 & 33.09 & 36.36 & 36.00 & 36.04 & 49.00 & 45.97 \\
  & Epoch 1.5        & 41.30&	39.84	&41.91&	40.93&	36.80&	38.54&	49.22&	44.74 \\
   & Epoch 2          & 41.84&	42.28	&38.24&	43.80&	44.80&	38.54&	53.12&	44.74\\

\bottomrule
\end{tabular*}
\end{table}

\begin{table*}[t]
\centering
\renewcommand{\arraystretch}{0.55}
\caption{Scores on \texttt{HalluSet-train} for models trained using GRPO with KL loss on QuALITY. \textbf{Adding the KL loss does not make the memorization effect vanish}.}

\label{tab:with_KL}
\scriptsize
\setlength{\tabcolsep}{3pt}
\resizebox{\textwidth}{!}{%
\begin{tabular}{cccccccccc}
\toprule
\multicolumn{10}{c}{\textbf{QuALITY -- Qwen3-4B}} \\
\midrule
\textbf{Eval ckpt} & $S_1$ (320) & $S_2$ (317) & $S_3$ (286) & $S_4$ (230) & $S_5$ (237) & $S_6$ (220) & $S_7$ (197) & $S_8$ (188) & Overall (1995) \\
\midrule
\setrange{66.52}{77.39}%
$M_0$ & 63.44 & 66.25 & 67.13 & 65.65 & 65.82 & 71.36 & 63.96 & 70.21 & \heat{66.52} \\
$M_1$ & \seen{\textbf{70.62}} & 67.82 & 67.48 & 68.26 & 71.73 & 71.82 & 65.48 & 65.96 & \heat{68.77} \\
$M_2$ & \seen{72.19} & \seen{\textbf{75.39}} & 72.38 & 70.00 & 72.57 & 77.27 & 70.56 & 69.68 & \heat{72.68} \\
$M_3$ & \seen{73.75} & \seen{80.44} & \seen{\textbf{75.87}} & 72.48 & 77.22 & 73.18 & 71.07 & 70.74 & \heat{74.77} \\
$M_4$ & \seen{71.88} & \seen{76.97} & \seen{77.62} & \seen{\textbf{75.22}} & 77.90 & 78.18 & 70.56 & 72.34 & \heat{75.22} \\
$M_5$ & \seen{74.06} & \seen{73.50} & \seen{76.22} & \seen{77.39} & \seen{\textbf{79.32}} & 77.91 & 73.60 & 74.47 & \heat{75.71} \\
$M_6$ & \seen{72.81} & \seen{75.08} & \seen{80.07} & \seen{78.26} & \seen{84.81} & \seen{\textbf{79.09}} & 74.62 & 78.19 & \heat{77.64} \\
$M_7$ & \seen{75.62} & \seen{76.34} & \seen{76.92} & \seen{77.39} & \seen{78.06} & \seen{79.55} & \seen{\textbf{75.63}} & 73.94 & \heat{76.69} \\
$M_8$ & \seen{73.44} & \seen{77.60} & \seen{77.97} & \seen{73.48} & \seen{81.86} & \seen{81.82} & \seen{78.17} & \seen{\textbf{76.06}} & \heat{77.39} \\
\midrule
\multicolumn{10}{c}{\textbf{QuALITY -- Qwen3-8B}} \\
\midrule
\textbf{Eval ckpt} & $S_1$ (295) & $S_2$ (303) & $S_3$ (243) & $S_4$ (206) & $S_5$ (199) & $S_6$ (192) & $S_7$ (162) & $S_8$ (140) & Overall (1740) \\
\midrule
\setrange{71.95}{82.25}%
$M_0$ & 73.90 & 74.26 & 68.72 & 69.90 & 70.85 & 72.40 & 70.37 & 74.29 & \heat{71.95} \\
$M_1$ & \seen{\textbf{75.59}} & 76.55 & 72.02 & 78.16 & 67.84 & 69.27 & 73.46 & 74.29 & \heat{73.68} \\
$M_2$ & \seen{77.29} & \seen{\textbf{80.86}} & 73.66 & 74.27 & 74.87 & 72.92 & 74.69 & 80.00 & \heat{76.26} \\
$M_3$ & \seen{80.68} & \seen{82.51} & \seen{\textbf{78.54}} & 76.16 & 80.90 & 78.12 & 73.46 & 72.86 & \heat{78.61} \\
$M_4$ & \seen{82.37} & \seen{83.83} & \seen{79.84} & \seen{\textbf{80.67}} & 80.70 & 78.65 & 77.78 & 78.57 & \heat{80.74} \\
$M_5$ & \seen{80.00} & \seen{84.16} & \seen{81.07} & \seen{81.55} & \seen{\textbf{83.92}} & 78.73 & 83.33 & 82.86 & \heat{81.90} \\
$M_6$ & \seen{81.69} & \seen{85.81} & \seen{79.84} & \seen{80.58} & \seen{84.92} & \seen{\textbf{82.29}} & 79.01 & 77.86 & \heat{81.91} \\
$M_7$ & \seen{82.71} & \seen{84.49} & \seen{80.25} & \seen{81.07} & \seen{82.41} & \seen{81.25} & \seen{\textbf{83.86}} & 77.14 & \heat{81.95} \\
$M_8$ & \seen{83.05} & \seen{82.18} & \seen{79.84} & \seen{80.58} & \seen{83.92} & \seen{82.29} & \seen{84.57} & \seen{\textbf{82.29}} & \heat{82.25} \\
\midrule
\multicolumn{10}{c}{\textbf{QuALITY -- Qwen3-14B}} \\
\midrule
\textbf{Eval ckpt} & $S_1$ (277) & $S_2$ (250) & $S_3$ (203) & $S_4$ (184) & $S_5$ (180) & $S_6$ (154) & $S_7$ (135) & $S_8$ (113) & Overall (1496) \\
\midrule
\setrange{77.90}{86.14}%
$M_0$ & 78.87 & 85.60 & 80.27 & 77.17 & 74.44 & 70.13 & 80.74 & 68.14 & \heat{77.90} \\
$M_1$ & \seen{\textbf{82.67}} & 80.40 & 82.22 & 78.80 & 72.22 & 72.08 & 79.26 & 73.45 & \heat{78.40} \\
$M_2$ & \seen{83.03} & \seen{\textbf{84.60}} & 82.73 & 80.98 & 81.67 & 72.73 & 82.96 & 78.76 & \heat{81.44} \\
$M_3$ & \seen{81.23} & \seen{86.80} & \seen{\textbf{83.74}} & 80.07 & 78.33 & 74.68 & 83.70 & 83.19 & \heat{81.71} \\
$M_4$ & \seen{87.73} & \seen{84.80} & \seen{85.71} & \seen{\textbf{82.61}} & 78.44 & 73.38 & 80.74 & 83.19 & \heat{82.77} \\
$M_5$ & \seen{83.75} & \seen{84.80} & \seen{87.19} & \seen{86.96} & \seen{\textbf{81.67}} & 77.92 & 80.74 & 82.30 & \heat{83.56} \\
$M_6$ & \seen{85.92} & \seen{84.40} & \seen{87.68} & \seen{83.15} & \seen{83.33} & \seen{\textbf{83.12}} & 83.93 & 82.30 & \heat{84.51} \\
$M_7$ & \seen{83.75} & \seen{85.60} & \seen{91.13} & \seen{89.13} & \seen{85.00} & \seen{80.52} & \seen{\textbf{88.15}} & 80.19 & \heat{85.68} \\
$M_8$ & \seen{86.28} & \seen{85.20} & \seen{88.67} & \seen{85.33} & \seen{84.22} & \seen{85.71} & \seen{89.63} & \seen{\textbf{84.19}} & \heat{86.14} \\
\midrule
\multicolumn{10}{c}{\textbf{QuALITY -- Qwen3-30B-A3B}} \\
\midrule
\textbf{Eval ckpt} & $S_1$ (274) & $S_2$ (282) & $S_3$ (205) & $S_4$ (176) & $S_5$ (150) & $S_6$ (134) & $S_7$ (134) & $S_8$ (118) & Overall (1473) \\
\midrule
\setrange{68.64}{80.99}%
$M_0$ & 68.25 & 65.60 & 71.71 & 71.02 & 76.00 & 61.94 & 67.91 & 66.95 & \heat{68.64} \\
$M_1$ & \seen{\textbf{70.80}} & 69.86 & 76.10 & 66.48 & 72.00 & 67.91 & 70.90 & 65.25 & \heat{70.26} \\
$M_2$ & \seen{71.90} & \seen{\textbf{74.47}} & 80.49 & 76.70 & 78.67 & 67.16 & 76.87 & 78.81 & \heat{75.42} \\
$M_3$ & \seen{77.37} & \seen{75.53} & \seen{\textbf{78.54}} & 75.57 & 82.00 & 71.64 & 77.61 & 74.58 & \heat{76.71} \\
$M_4$ & \seen{75.55} & \seen{79.79} & \seen{80.49} & \seen{\textbf{76.14}} & 86.00 & 73.13 & 78.36 & 74.58 & \heat{78.14} \\
$M_5$ & \seen{78.83} & \seen{80.50} & \seen{83.90} & \seen{80.68} & \seen{\textbf{86.00}} & 70.15 & 76.87 & 77.12 & \heat{79.70} \\
$M_6$ & \seen{78.83} & \seen{81.56} & \seen{85.37} & \seen{78.41} & \seen{83.33} & \seen{\textbf{72.39}} & 79.10 & 74.58 & \heat{79.77} \\
$M_7$ & \seen{77.37} & \seen{78.01} & \seen{82.93} & \seen{78.98} & \seen{86.00} & \seen{73.88} & \seen{\textbf{77.61}} & 75.42 & \heat{78.89} \\
$M_8$ & \seen{78.47} & \seen{81.56} & \seen{84.88} & \seen{78.98} & \seen{86.00} & \seen{76.12} & \seen{81.34} & \seen{\textbf{80.51}} & \heat{80.99} \\
\midrule
\multicolumn{10}{c}{\textbf{QuALITY -- Olmo-3-7B-Think}} \\
\midrule
\textbf{Eval ckpt} & $S_1$ (341) & $S_2$ (294) & $S_3$ (280) & $S_4$ (260) & $S_5$ (245) & $S_6$ (238) & $S_7$ (212) & $S_8$ (218) & Overall (2088) \\
\midrule
\setrange{76.11}{85.68}%
$M_0$ & 73.02 & 76.93 & 77.50 & 74.62 & 79.18 & 75.21 & 78.30 & 75.23 & \heat{76.11} \\
$M_1$ & \seen{\textbf{76.25}} & 77.93 & 80.36 & 78.46 & 77.55 & 78.99 & 76.89 & 75.69 & \heat{77.79} \\
$M_2$ & \seen{80.35} & \seen{\textbf{80.95}} & 80.00 & 81.54 & 80.41 & 77.31 & 75.94 & 78.44 & \heat{79.55} \\
$M_3$ & \seen{80.94} & \seen{81.29} & \seen{\textbf{83.21}} & 80.46 & 83.27 & 78.15 & 79.72 & 80.73 & \heat{81.05} \\
$M_4$ & \seen{83.58} & \seen{82.31} & \seen{83.57} & \seen{\textbf{83.85}} & 84.08 & 76.47 & 80.19 & 84.86 & \heat{82.47} \\
$M_5$ & \seen{85.04} & \seen{82.65} & \seen{87.50} & \seen{85.77} & \seen{\textbf{83.67}} & 80.25 & 80.38 & 81.19 & \heat{83.55} \\
$M_6$ & \seen{84.16} & \seen{85.37} & \seen{88.21} & \seen{82.69} & \seen{86.12} & \seen{\textbf{81.51}} & 81.96 & 84.86 & \heat{84.47} \\
$M_7$ & \seen{83.28} & \seen{85.03} & \seen{85.00} & \seen{85.38} & \seen{85.67} & \seen{81.09} & \seen{\textbf{85.49}} & 81.19 & \heat{84.06} \\
$M_8$ & \seen{85.92} & \seen{84.69} & \seen{88.57} & \seen{87.69} & \seen{86.12} & \seen{82.35} & \seen{85.55} & \seen{\textbf{83.82}} & \heat{85.68} \\
\midrule
\multicolumn{10}{c}{\textbf{QuALITY -- OLMoE-1B-7B}} \\
\midrule
\textbf{Eval ckpt} & $S_1$ (370) & $S_2$ (335) & $S_3$ (325) & $S_4$ (296) & $S_5$ (298) & $S_6$ (304) & $S_7$ (288) & $S_8$ (262) & Overall (2478) \\
\midrule
\setrange{56.54}{68.65}%
$M_0$ & 51.35 & 60.00 & 56.62 & 56.76 & 57.38 & 56.58 & 60.07 & 54.20 & \heat{56.54} \\
$M_1$ & \seen{\textbf{55.41}} & 60.90 & 52.92 & 60.81 & 57.72 & 60.86 & 57.29 & 56.11 & \heat{57.71} \\
$M_2$ & \seen{55.14} & \seen{\textbf{58.51}} & 54.77 & 60.81 & 60.40 & 57.89 & 57.99 & 57.25 & \heat{57.75} \\
$M_3$ & \seen{57.03} & \seen{61.49} & \seen{\textbf{59.38}} & 63.18 & 64.09 & 66.45 & 63.89 & 58.78 & \heat{61.66} \\
$M_4$ & \seen{57.30} & \seen{69.25} & \seen{61.23} & \seen{\textbf{63.51}} & 65.44 & 64.14 & 61.81 & 61.83 & \heat{62.99} \\
$M_5$ & \seen{58.92} & \seen{69.55} & \seen{62.77} & \seen{65.20} & \seen{\textbf{67.76}} & 60.49 & 67.71 & 64.12 & \heat{64.44} \\
$M_6$ & \seen{58.65} & \seen{63.58} & \seen{58.77} & \seen{65.88} & \seen{72.15} & \seen{\textbf{69.13}} & 64.24 & 66.41 & \heat{64.58} \\
$M_7$ & \seen{64.86} & \seen{72.54} & \seen{63.08} & \seen{71.28} & \seen{73.49} & \seen{69.41} & \seen{\textbf{71.88}} & 62.98 & \heat{68.65} \\
$M_8$ & \seen{63.51} & \seen{66.27} & \seen{61.23} & \seen{68.92} & \seen{70.13} & \seen{68.09} & \seen{70.14} & \seen{\textbf{66.91}} & \heat{66.72} \\
\bottomrule
\end{tabular}%
}
\end{table*}

\clearpage
\subsection{Prompts and examples}
\label{sec:prompts}

\newtcolorbox{promptbox}[1]{
  colback=black!5,     
  colframe=black!75,   
  fonttitle=\bfseries,
  title=#1,
  arc=2mm,             
  boxsep=5pt,
}

\begin{figure*}[ht]
\centering
\begin{minipage}{0.98\textwidth}
    \begin{promptbox}{Prompt for generating SFT data for QuALITY}
    \begin{Verbatim}[fontsize=\footnotesize, breaklines=true, breakanywhere=true, breaksymbolleft=]
You will be given a story, a multiple-choice question about it, and the correct answer to the question. Your task is to read the story carefully and explain clearly, in about four sentences, why the provided answer correctly answers the given question based on the story. In other words, you should provide the justification for why the given answer is correct.
Reason strictly based on the story. The explanation you provide should encapsulate the context required to correctly justify why the provided answer correctly answers the question. Do not add any extra text. Just provide the justification.
Do NOT produce any extra text referring back to the correct option number directly, i.e., do not include phrases like "Hence option 1 is the correct option."
    \end{Verbatim}
    \end{promptbox}
\end{minipage}
\caption{Prompt for generating SFT data for QuALITY.}
\label{fig:prompt_sft_data}
\end{figure*}

\begin{figure*}[ht]
\centering
\begin{minipage}{0.98\textwidth}
    \begin{promptbox}{Instruction template for post-training and evaluation for QuALITY data}
    \begin{Verbatim}[fontsize=\footnotesize, breaklines=true, breakanywhere=true, breaksymbolleft=]
You will be given a multiple-choice question about a story. Your goal is to provide the correct answer by carefully reasoning about the relevant information from the story. First, think about how to answer the question and then provide the answer. The thinking process should be enclosed within <think>...</think> tags. The final answer should be enclosed within <answer>...</answer> tags. An example of such a trajectory is: <think> [thinking process here] </think> <answer> [answer here]</answer>.
Answer the following question about the story {story_title} written by {story_author}.
    \end{Verbatim}
    \end{promptbox}
\end{minipage}
\caption{Instruction template for SFT, RL, and evaluations for QuALITY data.}
\label{fig:sft_template_QuALITY}
\end{figure*}

\begin{figure*}[t]
\centering
\begin{minipage}{0.98\textwidth}
    \begin{promptbox}{Instruction template for RL and evaluations for TriviaQA data}
    \begin{Verbatim}[fontsize=\footnotesize, breaklines=true, breakanywhere=true, breaksymbolleft=]
You will be given a trivia question. Your goal is to provide the correct answer by carefully reasoning about the relevant information. First, think about how to answer the question and then provide the answer. The thinking process should be enclosed within <think>...</think> tags. The final answer should be enclosed within <answer>...</answer> tags. An example of such a trajectory is: <think> [thinking process here] </think> <answer> [answer here] </answer>. Within the <answer></answer> tags, print only the correct answer to the question. Do not print anything else. Answer the following question: 
    \end{Verbatim}
    \end{promptbox}
\end{minipage}
\caption{Instruction template for RL and evaluations for TriviaQA data.}
\label{fig:sft_template_trivia}
\end{figure*}

\begin{figure*}[ht]
\centering
\begin{minipage}{0.98\textwidth}
    \begin{promptbox}{Example of a reasoning trace in QuALITY}
    \begin{Verbatim}[fontsize=\footnotesize, breaklines=true, breakanywhere=true, breaksymbolleft=]
Answer the following question about the story 'The Blue Behemoth' written by Brackett Leigh. Question: Why was Kapper in such a state of disbelief when Bucky and Jig found him? Option 1: He was frantically searching for the male Cansin he had found Option 2: He had lost all his animals and was desperate to find them Option 3: He had been attacked by the Vapor snakes Option 4: He was being poisoned. 
<think> Kapper was in a state of disbelief because he was bleeding, visibly sick, and incoherent, with jittery movements and bluish nails, which indicated poisoning. Later, the man who gave him the cigarette stood behind Jig, and it was implied that Kapper's drinks were spiked with "Selak," confirming he was being poisoned. The story reveals no direct connection between his state and searching for the male Cansin, losing animals, or being attacked by Vapor snakes, making poisoning the correct explanation.</think><answer> Correct option: 4. He was being poisoned.</answer>
    \end{Verbatim}
    \end{promptbox}
\end{minipage}
\caption{Example of a reasoning trace in QuALITY.}
\label{fig:quality_tt_example}
\end{figure*}

\begin{figure*}[ht]
\centering
\begin{minipage}{0.98\textwidth}
    \begin{promptbox}{Example of a reasoning trace in TriviaQA}
    \begin{Verbatim}[fontsize=\footnotesize, breaklines=true, breakanywhere=true, breaksymbolleft=]
 What is the name of the island that separates the two largest waterfalls at Niagara?
 <think>Okay, so the question is asking for the island that separates the two largest waterfalls at Niagara. Let me think. I know that Niagara Falls is actually a group of three waterfalls: the Horseshoe Falls, the American Falls, and the Bridal Veil Falls. But the two largest ones are probably the Horseshoe and the Bridal Veil Falls. No wait. Horseshoe and American Falls. Now, the island that separates them... I remember there's an island called Goat Island. Wait, but isn't there another one? Or maybe it's part of the split between the Canadian and American sides. Let me visualize the layout. The Horseshoe Falls are on the Canadian side, and the American Falls are on the U.S. side. The island that's in between them is Goat Island. But I also think there's an island called Luna Island, but I'm not sure. Wait, maybe Luna Island is part of the Niagara River near the falls but not exactly between the two main waterfalls. Wait, I think Goat Island is the one that splits the American Falls from the Horseshoe Falls. The American Falls are on the American side, and the Horseshoe is on the Canadian side. So Goat Island is in the middle, separating them. But I should confirm. Another thought: sometimes people refer to the island as part of the Niagara National Forests. Also, there's a bridge called the Niagara Falls Suspension Bridge that connects to Goat Island. I think that's right. So the answer is Goat Island.</think> <answer>Goat Island</answer>
    \end{Verbatim}
    \end{promptbox}
\end{minipage}
\caption{Example of a reasoning trace in TriviaQA.}
\label{fig:trivia_tt_example}
\end{figure*}

\begin{figure*}[ht]
\centering
\begin{minipage}{0.98\textwidth}
    \begin{promptbox}{Prompt for hallucination detection for QuALITY}
    \begin{Verbatim}[fontsize=\footnotesize, breaklines=true, breakanywhere=true, breaksymbolleft=]
You are a fact-checking assistant specializing in identifying hallucinated reasoning in LLM outputs. An LLM was given the following question about an article. The LLM first reasoned about the question and then answered it. Its reasoning may contain hallucinated (incorrect) facts not supported by the article.
Article: {article}
Question: {question}
LLM Reasoning: {model_reasoning}
LLM Answer: {model_answer}
Your task: Identify a fact in the reasoning that is clearly hallucinated - a fact that is NOT supported by the article but was central to how the LLM arrived at its answer.  If the reasoning chain quotes something as explicitly mentioned in the story while it is actually fully hallucinated, give this high weightage. Return only that fact that is totally hallucinated as one complete sentence, written without pronouns (use proper nouns/names instead so the sentence is self-contained). Return nothing else.
You must return just the hallucinated fact from the reasoning trace as it is (but as a complete sentence). Do not add justifications as to why it is hallucinated. If no part of the reasoning is hallucinated, return an empty string. The article, question, reasoning, and the LLM's answer are given below: 
    \end{Verbatim}
    \end{promptbox}
\end{minipage}
\caption{Prompt for hallucination detection for QuALITY.}
\label{fig:prompt_hall_detection_QuALITY}
\end{figure*}

\begin{figure*}[ht]
\centering
\begin{minipage}{0.98\textwidth}
    \begin{promptbox}{Prompt for generating MCQs from hallucinations for QuALITY}
    \begin{Verbatim}[fontsize=\footnotesize, breaklines=true, breakanywhere=true, breaksymbolleft=]
An LLM was given the following multiple-choice question, and it generated a reasoning and answer. The reasoning contains a hallucinated fact. Your task is to create a multiple-choice question that tests this hallucinated fact, i.e., the correct answer to your generated MCQ should be the hallucinated fact.

Original Question: {question}
LLM Reasoning: {model_reasoning}
LLM Answer: {model_answer}
Hallucinated Fact from the LLM Reasoning: {hallucinated_fact}

Instructions:
- Create one MCQ question that naturally combines the context of the original question and the hallucinated fact.
- The MCQ question you create should directly test the hallucinated fact provided to you. The question should NOT test the original question, its options, or the final answer generated by the model. 
- The correct answer should be the hallucinated fact (or a close paraphrase of it). 
- Add 3 plausible but wrong distractor options. None of the options should be "none of the above" or "all of the above".
- The question should sound like a standalone MCQ question, i.e., it should NOT start with or contain phrases like "According to the LLM's incorrect fact...", "According to the hallucinated detail...", etc. 
- Return only a valid Python dictionary with exactly these keys: testing_question, option1, option2, option3, option4, correct_answer.
- correct_answer must be only 1, 2, 3, or 4 (integer), indicating which option is correct.
--- FEW-SHOT EXAMPLES ---

    \end{Verbatim}
    \end{promptbox}
\end{minipage}
\caption{Prompt for generating MCQs from hallucinations for QuALITY.}
\label{fig:prompt_question_gen_QuALITY}
\end{figure*}

\begin{figure*}[ht]
\centering
\begin{minipage}{0.98\textwidth}
    \begin{promptbox}{Prompt for hallucination detection for TriviaQA}
    \begin{Verbatim}[fontsize=\footnotesize, breaklines=true, breakanywhere=true, breaksymbolleft=]
You are a fact-checking assistant specializing in identifying hallucinated reasoning in LLM outputs. An LLM was given the following trivia question. The LLM first reasoned about the question and then answered it. Its reasoning may contain hallucinated facts. A hallucinated fact is a statement that is factually incorrect according to world knowledge, i.e., it contradicts verifiable real-world information.
Your task: Identify 3 distinct hallucinated facts in the reasoning - facts that are NOT true but were central to how the LLM arrived at its answer or justified its answer. Each hallucinated fact must be a statement that is factually wrong according to world knowledge. Each fact must be written as a complete, self-contained sentence using proper nouns and names instead of pronouns, so it can be independently verified against the web. Do not add any justification or explanation.
Each identified hallucinated fact must be unique. Return a JSON object with exactly these keys: "f1", "f2", "f3". 
Question: {question}
LLM Reasoning: {model_reasoning}
LLM Answer: {model_answer}
 \end{Verbatim}
    \end{promptbox}
\end{minipage}
\caption{Prompt for hallucination detection for TriviaQA.}
\label{fig:prompt_hall_detection_trivia}
\end{figure*}

\begin{figure*}[ht]
\centering
\begin{minipage}{0.98\textwidth}
    \begin{promptbox}{Prompt for detecting backtracked hallucinations with TriviaQA}
    \begin{Verbatim}[fontsize=\footnotesize, breaklines=true, breakanywhere=true, breaksymbolleft=]

An LLM was given the following trivia question. The LLM first reasoned about the question and then answered it. While it answered the question correctly, its reasoning may contain hallucinated facts. A hallucinated fact is a statement that is factually incorrect according to world knowledge, i.e., it contradicts verifiable real-world information.

Question:
{question}

LLM Reasoning:
{model_reasoning}

LLM Answer:
{model_answer}

Your task: Identify 6 distinct atomic hallucinated facts in the reasoning - facts that are NOT true but were central to how the LLM arrived at its answer or justified its answer. Each fact must be written as a complete, self-contained sentence using proper nouns and names instead of pronouns, so it can be independently verified against the web. If the fact directly answers the question, it is likely not hallucinated since the final answer provided by the model is correct according to world knowledge. Each identified hallucinated fact must be unique.

The 6 facts should be split into two groups of 3, according to whether the model backtracked on the fact or not.

BACKTRACKED (3 facts): hallucinations where the model shows some backtracking or retracting behavior, i.e., facts that the model generates but regards as false later on. A backtracked fact is usually associated with alternate candidates where the model changes a proposed fact, in which case I want the first fact, i.e., the backtracked fact, to be chosen. If it says "Mars is the 1st planet in the Solar System. No, wait, it is the Earth.", I want you to pick "Mars is the 1st planet in the Solar System". 

UNBACKTRACKED (3 facts): hallucinations that are NOT associated with any backtracking behavior. The model must state the fact once and leave it standing: at no later point in the trace does it revise, correct, or replace the fact. 

The two groups must not overlap and must contain hallucinated facts. 

Return a JSON object with exactly these keys: "b_f1", "b_f2", "b_f3" for the backtracked facts, and "ub_f1", "ub_f2", "ub_f3" for the unbacktracked facts.

---FEW-SHOT EXAMPLES---

 \end{Verbatim}
    \end{promptbox}
\end{minipage}
\caption{Prompt for detecting backtracked hallucinations with TriviaQA.}
\label{fig:prompt_hall_detection_trivia_backtrack}
\end{figure*}

\begin{figure*}[ht]
\centering
\begin{minipage}{0.98\textwidth}
    \begin{promptbox}{Prompt for checking hallucinations using web search}
    \begin{Verbatim}[fontsize=\footnotesize, breaklines=true, breakanywhere=true, breaksymbolleft=]
You are a fact-checking assistant. You verify claims by searching the web. You are given three claims. Search the web to verify each one. For each claim, return 1 if the fact is true, and 0 if it is false. Base your evaluation on web searches from reliable sources. If a fact is partially true, extremely vague, or can not be verified, mark it as true. If it is definitely wrong, mark it as false. Return a JSON object with exactly these keys: "f1", "f2", "f3", where each value is 0 or 1. Return nothing else.
 \end{Verbatim}
    \end{promptbox}
\end{minipage}
\caption{Prompt for checking hallucinations using web search. We conservatively mark partially true, vague, or unverifiable claims as true to reduce false positives, since our goal is to identify clearly false factual claims as hallucinations.}\label{fig:prompt_web_check}
\end{figure*}

\begin{figure*}[ht]
\centering
\begin{minipage}{0.98\textwidth}
    \begin{promptbox}{Prompt for generating MCQs from hallucinations for TriviaQA}
    \begin{Verbatim}[fontsize=\footnotesize, breaklines=true, breakanywhere=true, breaksymbolleft=]
An LLM generated a hallucinated fact. A hallucinated fact is a fact that contradicts world knowledge and truth. Your task is to create a multiple-choice question that tests this hallucinated fact, i.e., the correct answer to your generated MCQ should be the hallucinated fact.

Hallucinated Fact produced by the LLM: {hallucinated_fact}

Instructions:
- Create one MCQ question that directly tests the given hallucinated (incorrect) fact.
- The correct answer to the question you frame should be the hallucinated fact (or a close paraphrase of it). 
- Add 3 plausible but wrong distractor options. None of the options should be "none of the above" or "all of the above".
- Avoid using negations while phrasing any question or any option, i.e., avoid words or constructions such as 'not'.
- The question should sound like a standalone MCQ question, i.e., it should NOT start with or contain phrases like "According to the LLM's incorrect fact...", "According to the hallucinated detail...", etc. 
- Return ONLY a valid Python dictionary with exactly these keys: testing_question, option1, option2, option3, option4, correct_answer.
- correct_answer must be only 1, 2, 3, or 4 (integer), indicating which option is correct.
- No explanation, no extra text - just the dictionary.
--- FEW-SHOT EXAMPLES ---

 \end{Verbatim}
    \end{promptbox}
\end{minipage}
\caption{Prompt for generating MCQs from hallucinations for TriviaQA.}
\label{fig:prompt_question_gen_trivia}
\end{figure*}

\begin{figure*}[ht]
\centering
\begin{minipage}{0.98\textwidth}
    \begin{promptbox}{Prompt for identifying all hallucinations on QuALITY data}
    \begin{Verbatim}[fontsize=\footnotesize, breaklines=true, breakanywhere=true, breaksymbolleft=]
You are a fact-checking assistant specializing in identifying hallucinated reasoning in LLM outputs. An LLM was given the following question about an article. The LLM first reasoned about the question and then answered it. Its reasoning may contain hallucinated (incorrect) facts not supported by the article.
Article: {article}
Question: {question}
LLM Reasoning: {model_reasoning}
LLM Answer: {model_answer}

Your task: Split the LLM Reasoning into individual sentences. For each sentence, decide whether it is a hallucination (a fact that is NOT supported by the article) or not. Label each sentence with 0 if it is hallucinated, 1 if it is not hallucinated, and 2 if it contains no factual information at all. Return ONLY a valid JSON object where each key is a sentence from the reasoning (verbatim, as a complete sentence) and each value is 0, 1, or 2. Do not add any justifications, explanations, markdown, or other text. Return nothing else.
 \end{Verbatim}
    \end{promptbox}
\end{minipage}
\caption{Prompt for identifying hallucinations in reasoning trace sentences using GPT-5.1-mini for QuALITY. We use an identical prompt for TriviaQA, but we do not use web search here, as running it over every sentence of a long reasoning trace is expensive.}
\label{fig:prompt_sentence_fact_check}
\end{figure*}

\end{document}